\documentclass[]{fairmeta}
\usepackage[utf8]{inputenc}
\usepackage{textcomp}
\usepackage{lmodern}

\usepackage{amsmath,amsfonts,bm}

\def\eqref#1{equation~\ref{#1}}

\def\1{\bm{1}}

\DeclareMathAlphabet{\mathsfit}{\encodingdefault}{\sfdefault}{m}{sl}
\SetMathAlphabet{\mathsfit}{bold}{\encodingdefault}{\sfdefault}{bx}{n}

\usepackage{amsmath,amssymb,mathtools}
\usepackage{booktabs,tabularx,array,multirow,makecell}
\usepackage{graphicx}
\usepackage{wrapfig}
\usepackage{xcolor}
\usepackage{colortbl}
\definecolor{tablehighlight}{HTML}{E8DFF2}
\definecolor{tableablation}{HTML}{F7F5FA}
\definecolor{tableaccent}{HTML}{61477C}
\definecolor{tablerule}{HTML}{A89CB8}
\newcommand{\tableheading}[1]{\textbf{#1}}
\newcommand{\tablemethod}[1]{\textcolor{tableaccent}{\textbf{#1}}}
\usepackage{float}
\usepackage{flafter}
\usepackage{placeins}
\usepackage{tikz}
\usetikzlibrary{arrows.meta,positioning,fit,calc,shapes.geometric,plotmarks}
\usepackage{microtype}
\usepackage{hyperref}
\usepackage{url}
\usepackage{environ}
\renewcommand{\arraystretch}{1.12}
\setlabdisplayname{OmniAI Group of ZJU ACES Lab}
\setuniversityname{}
\setdocumentlabel{Preprint}
\title{\fontsize{17}{21}\selectfont Recover, Don't Just Continue: Learning from Feedback for Physical Failure Understanding and Correction}
\author[1,*]{Miao Pan} \author[1,*]{Kaixiang Yao} \author[1]{Xiaohe Zhou} \author[1]{Xu Wang} \author[1]{Qinfeng Li} \author[1]{Wenqi Zhang} \author[1]{Xuhong~Zhang} \affiliation[1]{Zhejiang University} \contribution[*]{Equal contribution} \metadata[Project]{\url{https://pm1255.github.io/embodied-corrector/}} \metadata[Email]{\href{mailto:pan_136166@163.com}{\texttt{pan\_136166@163.com}}}

\newcommand{\embodiedabstracttext}{}
\RenewEnviron{abstract}{\global\let\embodiedabstracttext\BODY}
\begin{abstract}
Multimodal large language models are increasingly regarded as the cognitive foundation of embodied intelligence, yet planning from instructions alone does not make them reliable embodied brains. During physical execution, actions deviate from intended effects, and what can be realized is jointly constrained by the embodiment, controller, and environment. An embodied brain must therefore understand why execution deviates, identify what the current execution system can reliably realize, and revise decisions using interaction experience. These three requirements correspond to causal failure localization, capability-constrained adaptation, and experience-driven revision, which are complementary and non-substitutable. Although current embodied foundation models have advanced spatio-temporal reasoning, physical-world modeling, and high-level planning, they still largely lack these capabilities. To address this gap, we introduce Embodied-Corrector. We construct controlled failure–recovery data by injecting a single controlled perturbation at subtask boundaries of successful trajectories (navigation, reachability, grasp pose, control timing, articulation state, task prerequisite, or target grounding), replaying the affected subtask, and retaining only failures that are observable from public execution evidence and recoverable from the current state; we then align actions, observable consequences, failure causes, and executable recoveries into Judge–Cause–Sub–Fine supervision. Training uses evaluation-first SFT: the model first learns Judge and Cause, and then learns Sub and Fine conditioned on accurate Judge/Cause prefixes. Because SFT uses accurate prefixes while deployment can only use self-generated prefixes, evaluation errors propagate into recovery planning, and response-level rewards are too coarse to locate the affected planning tokens. We therefore use OPD to internalize the planning distribution under accurate Judge/Cause prefixes into the planning distribution under self-generated prefixes: during training, the model generates plans from self-generated Judge/Cause prefixes, while a contrastive forward pass with accurate Judge/Cause prefixes compares token probabilities and reinforces or suppresses planning tokens; at deployment, this contrastive branch is removed and no labels are exposed. Experiments on embodied failure reasoning and planning benchmarks show stronger failure-cause analysis and corrective planning: with a frozen VLA, composite task success improves by 6.1 percentage points, and with fixed GPT-5.5 execution, embodied manipulation and navigation tasks improve by 13.9 percentage points.
\end{abstract}

\def\abstractlist{{\color{metafg}\embodiedabstracttext}}
\begin{document}
\thispagestyle{firstheader}
\maketitle
\pagestyle{empty}
\section{Introduction}
\label{sec:introduction}

Multimodal large language models are becoming an important foundation for embodied intelligence~\citep{palm-e,rt2,saycan,gato}. By combining visual understanding with language reasoning, they can interpret task goals, reason about scene relations, and decompose complex instructions into executable steps~\citep{codeaspolicies,innermonologue,progprompt,llm-planner,pan2026groundwhatyousee}. Recent embodied foundation models such as RynnBrain~\citep{rynnbrain2026}, RoboBrain~2.0~\citep{robobrain2026}, and Embodied-Reasoner~\citep{embodiedreasoner2026} have advanced spatio-temporal memory, physical-world reasoning, and embodied planning; VeBrain-1.5~\citep{vebrain15} further unifies multimodal perception, spatial reasoning, and robot control in a shared decision interface. However, a reliable embodied brain requires more than planning. During physical execution, actions deviate from intended effects, and what can be realized is jointly constrained by the embodiment, controller, and environment~\citep{worldactionmodels2026}. An embodied brain must therefore understand why execution deviated, recognize what the current executor can reliably realize, and adjust decisions using interaction experience. This raises the central question: \emph{how can a multimodal model evolve from planning what to do into an embodied brain that understands physical interaction, adapts to execution capability, and corrects its behavior after failure?}

We argue that this requires three related but non-substitutable capabilities. First, \textbf{physical causal understanding and failure localization}~\citep{robofailring2026,neurvla2026,elfvla2026} requires determining not only that an execution failed, but also which action, physical relation, or unmet prerequisite caused it. Existing failure-memory and detection methods can flag unsuccessful outcomes~\citep{robofailring2026}, yet a model may observe an object not placed into a container while failing to distinguish incorrect grasp timing, inappropriate contact pose, or an earlier uncompleted step. Effective correction requires tracing execution back to the condition that first invalidated progress. Second, \textbf{execution-capability identification and adaptation}~\citep{pi05survey2026,octo} requires translating intent into a form the current executor can reliably realize. Even a correct diagnosis is ineffective if the recovery exceeds the controller's capability. The brain should adapt instruction content and granularity, e.g., decomposing a long request into state-conditioned subtasks for a frozen VLA~\cite{pi05survey2026,rt1,rtx}. Third, \textbf{in-context learning from interaction experience}~\citep{zeva2026,retrievalvla2026,ricl2025} requires information from one execution to change the next decision. Prior work indicates that a robot's own physical interaction experience can support in-context learning even when the policy model is frozen~\citep{zeva2026}. Failure observations, previous actions, and successful corrections should become evidence for planning rather than accumulating in context while the model repeats the same behavior~\citep{embodiedfailure2026,learningtrials2026}.

\begin{figure}[t]
    \centering
    \includegraphics[width=0.78\linewidth]{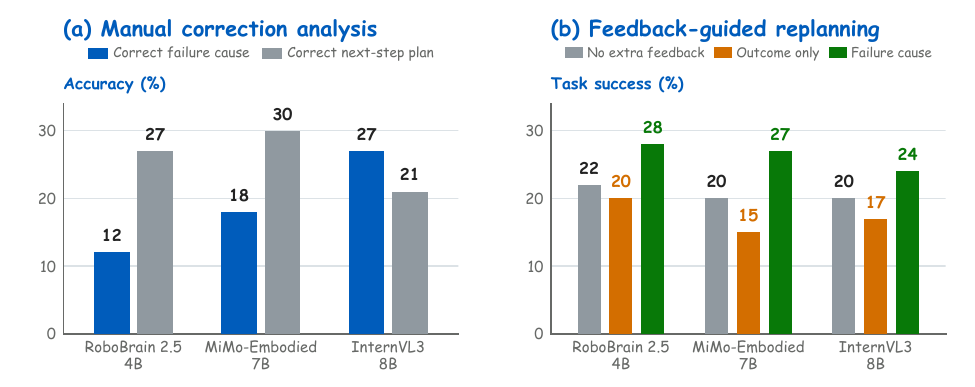}
    \caption{Failure diagnosis and feedback controls on EB-ALFRED. (a) Frozen multimodal models struggle to identify the physical cause of failed execution. (b) With the model and action interface fixed, cause-level feedback improves replanning more than repeating the failed outcome.}
    \label{fig:motivation_feedback}
\end{figure}

We first examine whether current embodied brains possess these capabilities. On failed EB-ALFRED~\citep{embodiedbench} executions, we supply frozen multimodal models with the same task goal, visual observations, executed actions, and execution feedback, and inspect predicted failure cause and proposed next-step correction. As shown in Figure~\ref{fig:motivation_feedback}(a), models such as RynnBrain~\citep{rynnbrain2026}, MiMo-Embodied~\citep{mimoembodied2026}, and InternVL3~\citep{internvl3} frequently produce plausible descriptions while failing to identify the condition that actually blocks progress, and their plans often fail to repair it. This exposes weak physical causal understanding and failure localization. The missing capability is not merely recognizing failure, but transforming execution evidence into an explanation of \emph{why} it failed and \emph{what must change next}.

We next ask what form this missing information should take. Keeping the evaluated brain, visual input, action interface, and decoding fixed on EB-ALFRED, we vary only the feedback after failure. As shown in Figure~\ref{fig:motivation_feedback}(b), repeating the failed outcome provides little benefit~\citep{reflexion}, whereas supplying the concrete failure cause consistently improves performance. For RoboBrain2.5-4B, MiMo-Embodied-7B, and InternVL3-8B, cause-level feedback changes task success from $0.22$ to $0.28$, from $0.20$ to $0.27$, and from $0.20$ to $0.24$. Useful feedback does not simply state failure; it identifies which expected condition was violated and why that prevents progress, revising the brain's interpretation of the task state. Together, these diagnostics indicate that current models lack not only failure analysis, but also execution-capability adaptation and in-context learning from interaction: outcome repetition alone does not change subsequent decisions, and only explicit causal feedback enables correction.

Motivated by this, we introduce \textbf{Embodied-Corrector}, which couples the three capabilities through data, training, and inference. For \emph{physical causal understanding and failure localization}, we build controlled failure--recovery data by branching successful trajectories at subtask boundaries and intervening on navigation, reachability, grasp pose, control timing, articulation state, task prerequisites, and target grounding. Each branch shares the preceding successful context while changing one controlled factor, and we retain only failures observable from public execution evidence and recoverable from the realized state. The resulting paired failure--recovery experience connects what changed, what consequence it caused, and what must be restored.

To obtain such data at scale, we build an automated pipeline that coordinates simulation environments, robot execution interfaces, multimodal models, and tools for trajectory generation, failure construction, cause annotation, and correction verification, with a Harness providing runtime support for interaction and data collection. Each sample contains four fields: Judge determines whether the current subtask is completed; Cause identifies the failure cause supported by observable evidence; Sub decides whether to advance to the next subtask or redo the current subtask; and Fine specifies how to execute Sub, including actions, target arguments, execution order, and verification steps. Learning from such data is challenging: Judge/Cause errors contaminate Sub/Fine; SFT trains Sub/Fine with accurate Judge/Cause prefixes, whereas deployment can only use self-generated Judge/Cause, creating a prefix distribution mismatch; response-level rewards are too coarse to locate the planning tokens affected by evaluation errors; and Fine must satisfy the realizability constraints of a frozen executor. We therefore adopt “evaluation-first SFT + OPD internalization.” In the first stage, we train Judge/Cause, and then supervise Sub/Fine generation with accurate Judge/Cause prefixes. In the second stage, we use OPD: during training, the model generates Sub/Fine from self-generated Judge/Cause prefixes, while a contrastive forward pass with accurate Judge/Cause prefixes compares token probabilities and reinforces planning words supported by accurate diagnoses or suppresses contradictory ones. Accurate diagnoses are used only for scoring and do not replace self-generated prefixes; at deployment, the contrastive branch is removed, and the model still autonomously generates Judge/Cause and Sub/Fine. Thus, a single inference outputs the complete Judge–Cause–Sub–Fine and produces state-conditioned recovery plans realizable by a frozen executor.

Experiments evaluate corrective reasoning and execution with fixed backends. First, on brain-level corrective reasoning, Embodied-Corrector substantially improves failure-cause analysis and produces recovery plans consistent with the identified physical cause, demonstrating stronger physical causal reasoning rather than outcome classification alone. Second, with the VLA frozen, execution-aware decomposition improves $\pi_{0.5}$ by \textbf{5.4 percentage points} on Composite-Seen tasks and by \textbf{6.1 percentage points} on Composite-Unseen tasks, showing that the brain can adapt task granularity to reduce the burden on a fixed executor. Third, with the GPT-5.5 code generator, robot functions, perception module, and execution budget all fixed, feeding Embodied-Corrector's diagnosis and recovery back into the next planning context improves all ten execution conditions over the untrained corrective brain, with a macro-average gain of \textbf{13.9 percentage points}. This shows that execution feedback is not merely stored in history: it changes the next workflow produced by the same planner. Our contributions are threefold:
\begin{itemize}
    \item We identify physical causal understanding, execution-capability adaptation, and in-context learning from interaction feedback as three complementary capabilities for corrective embodied reasoning, and diagnose their absence in current multimodal embodied brains.
    \item We introduce Embodied-Corrector and a controlled failure--recovery data construction pipeline connecting actions, observable physical consequences, failure causes, and executable recoveries, together with evaluation-first SFT.
    \item We evaluate correction from brain-level failure reasoning to fixed-backend execution, showing the learned brain can explain failures, adapt instructions to a frozen VLA, and provide corrective context that changes the subsequent thinking of a code generator.
\end{itemize}

\section{Related Work}
\paragraph{Embodied agent architectures.}
Language-model robot systems commonly separate semantic planning from physical execution~\citep{wang2026thea,galanti2026physicalagency,liu2026phyagentos}. SayCan grounds skill
selection with learned affordances, Inner Monologue returns environment feedback to the planning context, and
ReAct alternates reasoning with tool use~\citep{harness_ahn2022saycan,harness_huang2022inner,harness_yao2022react}.
Code as Policies turns language plans into executable programs, while CaP-X evaluates code agents through
perception, geometry, and manipulation interfaces~\citep{harness_liang2023code,fu2026capx}. Harness VLA routes
plans to frozen VLA skills, and FineVLA and InstructVLA show that language guidance must retain execution-critical
attributes~\citep{zhang2026harnessvla,hu2026finevla,yang2026instructvla,feng2026embodied,yao2026spatial}. These systems motivate our fixed
execution backends, but our central question is upstream: how failure evidence should change the brain plan before
the backend attempts the next recovery.

\paragraph{Embodied execution correction.}
Correction begins with feedback grounded in what the robot actually did~\citep{huang2026maskharness}. Inner Monologue feeds success, scene, and
human feedback back into high-level planning, while model-based runtime monitoring detects anomalous or risky
execution states~\citep{harness_huang2022inner,liu2023runtime}. At the controller level, VLA-Corrector compares
predicted and observed visual dynamics, truncates stale action chunks, and triggers local replanning
~\citep{pan2026vlacorrector}. Language-agent methods such as Reflexion and Self-Refine revise a prior answer from
textual feedback~\citep{shinn2023reflexion,harness_madaan2023selfrefine}, but their feedback need not identify a
physically observable cause or verify that a repair changes the environment. ASPIRE comes closest to that closed
loop: its execution engine exposes fine-grained multimodal traces for failure diagnosis, repair synthesis, and
validation~\citep{lu2026aspire}. Embodied-Reasoner, RynnBrain, and RxBrain additionally provide multimodal task
reasoning and execution evaluation~\citep{zhang2025embodiedreasoner,dang2026rynnbrain,tencent2026rxbrain}. Our focus
is the missing task-level connection among these capabilities: after each completed subtask, identify the observable
cause, revise the remaining dependency plan from the current state, and verify that the same fixed executor realizes
the correction.

\paragraph{Memory for embodied agents.}
Voyager stores reusable skills; BrainMem separates working, episodic, and semantic memory; RoboMemory retrieves procedural experience; and Harness VLA retains execution traces
\citep{wang2023voyager,wang2026brainmem,wang2025robomemory,zhang2026harnessvla}. These systems show that memory
supports long-horizon state tracking, similar-task planning, and skill reuse, but retrieval alone does not show that a repeated physical error was prevented \citep{liang2026embodiedmemory}. Our architecture therefore separates volatile working state from environment-validated correction episodes and only consolidates a semantic rule after multiple compatible successes.
This write policy treats memory as evidence-backed context for correction rather than as authority over the current
observation.
\section{Failure Data and Corrective Brain Training}
\label{sec:method}

Recent code-based embodied systems improve robot-function invocation, skill composition, and routing to fixed execution backends, while execution-aware frameworks provide trajectory feedback for diagnosis and recovery~\citep{fu2026capx,ding2026zeta,pan2026vlacorrector}. However, they provide limited supervision for task-level failure understanding: models are rarely trained to judge whether a subtask succeeded, identify the broken observable prerequisite, and decide whether to advance or recover. We therefore construct controlled failure trajectories from successful demonstrations and convert them into supervision for outcome judgment, cause diagnosis, recovery-subtask selection, and fine-grained planning. Because evaluation and planning are difficult to learn jointly, we use an evaluation-first curriculum and finally apply on-policy distillation (OPD) to internalize verified evaluation into the planning-token distribution.

\subsection{Dataset and Benchmark Construction}
\label{sec:data_construction}

\begin{figure*}[!htbp]
    \centering
    \includegraphics[width=1\linewidth]
    {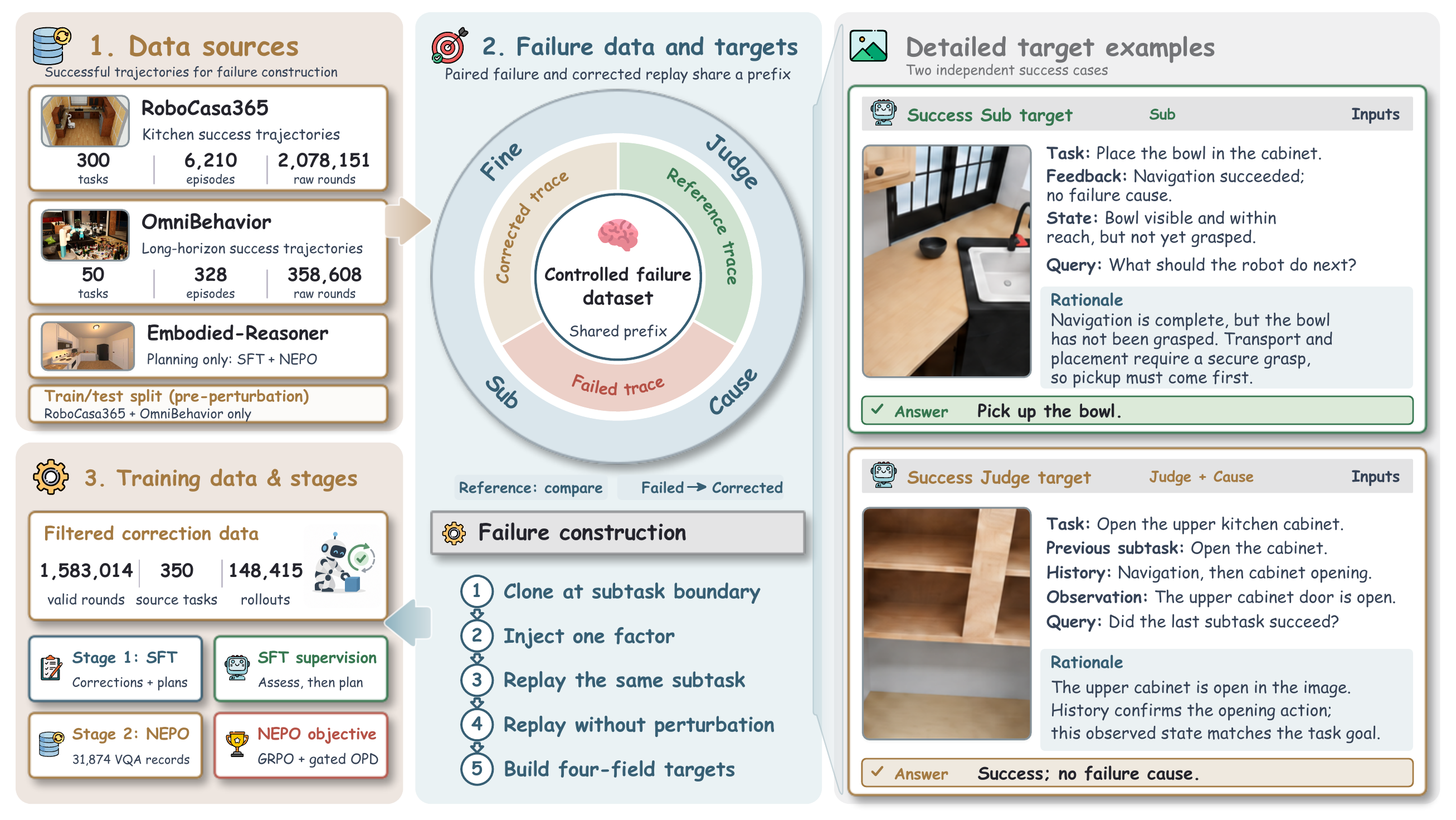}
    \caption{Controlled failure-data construction.
Successful trajectories are branched through controlled perturbations and replayed to produce aligned supervision for outcome judgment, cause diagnosis, recovery-subtask selection, and fine-grained planning.}
    \label{fig:dataset_full}
\end{figure*}

\paragraph{Controlled failure trajectories.}
We begin with successful RoboCasa365 and OmniBehavior trajectories~\citep{nasiriany2026robocasa365,li2024behavior1k}. Subtask boundaries are identified from task-graph transitions and task-relevant state changes, such as reaching a target, grasping an object, or opening a container. At each boundary, we retain the task goal $g_i$, current subtask, executed workflow $w_i$, pre-execution observation $o_i$, and corresponding successful segment. We then clone the simulator state before the subtask, inject one known perturbation, and execute the same subtask again. Each branch changes only one factor, making its failure source identifiable from the intervention. Perturbations cover navigation, reachability, grasp pose, control timing, articulation, task prerequisites, and target grounding. The replay produces a post-execution observation $o_{i+1}$ and execution trace $\tau_i$, including function arguments, returns, exceptions, and key visual frames.

\paragraph{Correction-target construction.}
Judge and Cause are generated from the known perturbation and intervention point. Judge records whether the perturbed execution completes the current subtask. Cause combines the perturbation with its realized consequence and describes the failure using publicly observable evidence. For example, a changed grasp height that leaves the object unattached is written as ``the gripper closed above the target and the object was not grasped,'' rather than exposing a hidden simulator parameter. For a failed example, the current subtask remains incomplete and is used as Sub. For a successful example, Sub is the next subtask in the original trajectory. Thus, Sub measures whether the model can decide to advance or repeat the interrupted step. Fine specifies how to execute Sub. We replay a corrective execution from the realized failed state to obtain an aligned corrected segment. A fixed closed-source model summarizes the failed and corrected traces into an ordered fine-grained plan while preserving the target object, motion direction, end-effector state, action order, and verification steps. Fine is therefore a language description of the corrected trajectory rather than a template selected from the perturbation type.

\paragraph{Filtering and benchmark construction.}
We discard ineffective perturbations, unsynchronized media, inconsistent object identities, invalid traces, and failures that cannot be identified from observations. Source trajectories are split before perturbation generation. The corrective training set contains aligned Judge, Cause, Sub, and Fine targets from training trajectories. \textbf{EmbodiedDiag-Bench} is generated only from held-out trajectories and therefore shares neither source trajectories nor successful prefixes with training. We report each field separately, while the joint score measures whether the model can transform execution evidence into a corrected task state and recovery decision.

\subsection{Evaluation-First Brain Training and OPD Internalization}
\label{sec:training}

\begin{figure*}[!htbp]
\centering
\includegraphics[width=\linewidth]{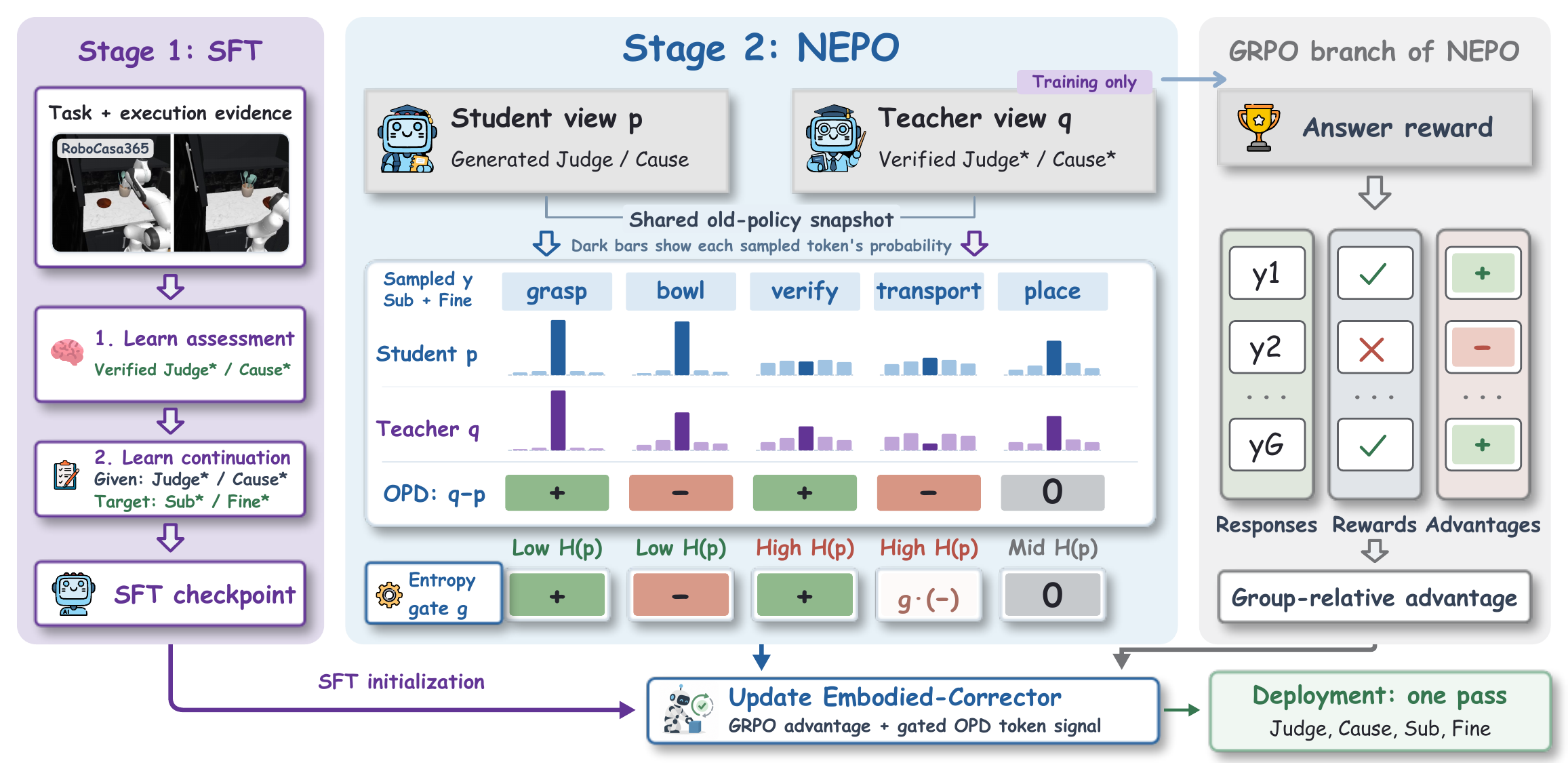}
\caption{Overview of Embodied-Corrector training.
Training proceeds from execution evaluation to recovery planning, followed by NEPO optimization that reduces the mismatch between verified evaluation prefixes used during training and model-generated prefixes used at deployment.}
\label{fig:brain_training}
\end{figure*}

Embodied-Corrector is a unified multimodal VQA model that autoregressively generates Judge, Cause, Sub, and Fine. An incorrect evaluation prefix can distort the recovery plan, while a response-level reward is too coarse to identify the responsible planning token. We therefore first train evaluation, then recovery planning, and finally use OPD to internalize verified evaluation. Deployment still uses one inference to generate the complete response.

\paragraph{Stage 1: Evaluation-first supervised training.}
We first train Judge and Cause, requiring the model to determine whether the subtask was completed and explain its observable failure cause. We then train Sub and Fine conditioned on verified Judge and Cause. This curriculum teaches the dependency from execution outcome to cause diagnosis and recovery planning without requiring difficult joint prediction from the beginning.

\paragraph{Stage 2: OPD-based evaluation internalization.}
Teacher-forced training generates Sub and Fine conditioned on accurate Judge/Cause, whereas deployment uses the model's own generated Judge/Cause; this prefix distribution mismatch allows evaluation errors to propagate into subsequent planning tokens. We therefore use OPD to internalize the planning distribution under accurate Judge/Cause prefixes into the planning distribution under the model's own prefixes. For the \(i\)-th training sample, let \(x_i\) denote the input, including the task goal, executed workflow, pre- and post-execution observations, and execution trace; let \(a_i^*=(d_i^*,c_i^*)\) denote the accurate Judge/Cause, where \(d_i^*\in\{0,1\}\) is the subtask-completion label and \(c_i^*\) is the failure-cause label obtained from controlled perturbation and replay; let \(\hat{a}_i=(\hat{d}_i,\hat{c}_i)\) denote the model's self-generated Judge/Cause; let \(y_i=(y_{i,1},\ldots,y_{i,T})\) denote the planning token sequence for Sub/Fine, where \(T\) is the sequence length, \(t\in\{1,\ldots,T\}\) is a planning-token position, \(v\) is a candidate token in the vocabulary, \(y_{i,<t}\) denotes the first \(t-1\) generated planning tokens, and \(\pi_{\theta_{\mathrm{old}}}\) is the old-policy snapshot before the update, used to compute both prefix distributions under the same policy for comparability. Using the same old policy, we compute the probability of token \(v\) at the \(t\)-th planning position under the two prefixes:
\begin{equation}
p_{i,t}(v)=\pi_{\theta_{\mathrm{old}}}\left(v\mid x_i,\hat{a}_i,y_{i,<t}\right),\qquad
q_{i,t}(v)=\pi_{\theta_{\mathrm{old}}}\left(v\mid x_i,a_i^*,y_{i,<t}\right),
\end{equation}
where \(p_{i,t}\) is the distribution under the model's self-generated Judge/Cause prefix and \(q_{i,t}\) is the distribution under the accurate Judge/Cause prefix; their token-level signed advantage is
\begin{equation}
s_{i,t}^{\mathrm{OPD}}(v)=q_{i,t}(v)-p_{i,t}(v).
\end{equation}
When \(s_{i,t}^{\mathrm{OPD}}(v)>0\), the accurate Judge/Cause increases support for token \(v\), so this token should be reinforced; when \(s_{i,t}^{\mathrm{OPD}}(v)<0\), the accurate Judge/Cause decreases support for token \(v\), so this token should be suppressed. Note that \(a_i^*\) is used only to compute \(q_{i,t}\) and score planning tokens, and does not replace the model's self-generated prefix \(\hat{a}_i\); at deployment, the accurate branch is removed entirely and no labels are exposed.

\paragraph{Noise-aware asymmetric OPD.}
Under prefix mismatch, \(p_{i,t}\) at high-entropy positions is flatter, making \(s_{i,t}^{\mathrm{OPD}}\) more sensitive to prefix noise; negative signals may conflict with the task-level RL reward and destabilize training~\citep{shao2024deepseekmath,zhong2026vlaopd,lu2026selfdistilled}. We therefore introduce NEPO: for the \(t\)-th planning position, we first compute the entropy of the self-generated distribution \(p_{i,t}\),
\begin{equation}
h_{i,t}=-\sum_{v}p_{i,t}(v)\log p_{i,t}(v),
\end{equation}
and then define the gate value
\begin{equation}
g_{i,t}=\sigma\left(\frac{\kappa-h_{i,t}}{\beta}\right),
\end{equation}
where \(\sigma\) is the sigmoid function, \(\kappa\) is the entropy threshold, and \(\beta>0\) is the gate temperature. The gate decreases smoothly with entropy, attenuating negative signals more strongly at uncertain positions. Evaluating \(s_{i,t}^{\mathrm{OPD}}\) at the sampled token \(y_{i,t}\), the final token-level advantage is
\begin{equation}
\delta_{i,t}^{\mathrm{NEPO}}=\left[s_{i,t}^{\mathrm{OPD}}\right]_{+}+g_{i,t}\left[s_{i,t}^{\mathrm{OPD}}\right]_{-},
\end{equation}
where \([s]_{+}=\max(s,0)\) retains positive signals and \([s]_{-}=\min(s,0)\) retains negative signals; that is, positive advantages are always retained, while negative advantages are attenuated according to entropy. The complete response still receives the task-level VQA/RL reward \(r\), and each planning token additionally receives \(\lambda_{\mathrm{OPD}}\delta_{i,t}^{\mathrm{NEPO}}\), where \(\lambda_{\mathrm{OPD}}>0\) controls the relative weight of the OPD term; the two are jointly updated under a GRPO-style objective, and only Embodied-Corrector is updated. At deployment, the accurate Judge/Cause branch is removed, and a single inference generates the complete Judge--Cause--Sub--Fine response.
\subsection{Corrective Harness}
\label{sec:harness_overview}

\begin{figure}[htbp]
\centering
\includegraphics[width=0.98\linewidth]{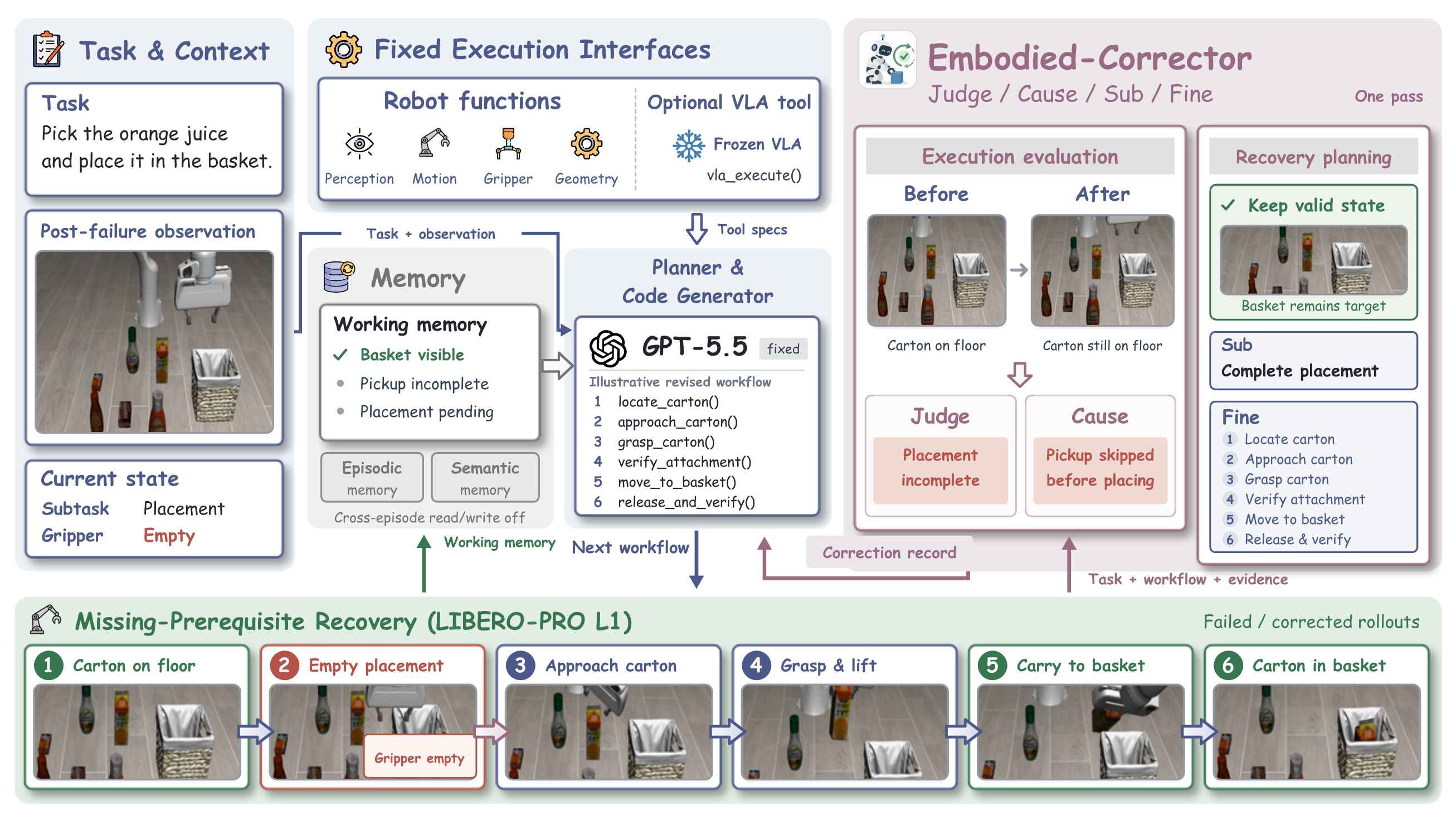}
\caption{Architecture and closed-loop workflow of the Embodied-Corrector Harness. Separating execution evaluation, recovery planning, fixed code/VLA realization, and environment-validated memory turns failure evidence into executable correction while preserving prior progress.}
\label{fig:harness_architecture}
\end{figure}

Figure~\ref{fig:harness_architecture} shows how the trained brain enters the execution loop. A fixed closed-source planner combines the goal, current observation, execution history, and retrieved memory to generate workflow code over analytic robot functions or a frozen VLA. After execution, Embodied-Corrector reads the goal, executed workflow, boundary observations, and public trace, then returns Judge, Cause, Sub, and Fine. The planner uses this record to revise the next workflow from the current state while retaining completed progress. Retrieved memory is provided to the planner; corrections enter episodic memory only after environment verification. Appendix~\ref{sec:harness} gives the formal interface, memory policy, and detailed workflow.

\section{Experiments}
\label{sec:experiments}

We first evaluate execution understanding and corrective planning on EmbodiedDiag-Bench, followed by transfer to EB-ALFRED. Harness-level experiments then test two mutually exclusive simulated execution settings: code workflows with the VLA function disabled, and VLA-only workflows with all analytic robot functions disabled. We next analyze post-training dynamics and corrective judgments as subtask rewards, and conclude with physical validation on the Piper platform.

\subsection{Brain-Level Evaluation}
\label{sec:brain_eval}
\paragraph{Corrective reasoning on EmbodiedDiag-Bench.}
We first remove the code and action executors and directly test whether the brain can produce a coherent correction from a completed physical execution. EmbodiedDiag-Bench contains held-out navigation and manipulation failures and evaluates four fields. Judge determines whether the current subtask was completed, Cause identifies the failed entity, relation, and outcome, Sub decides whether to advance or complete the current subtask again, and Fine specifies the required actions, grounded arguments, ordering, and intermediate checks.

\begin{table*}[!htbp]
\caption{Brain-level corrective reasoning on EmbodiedDiag-Bench
(\%). Judge and Cause measure whether the brain understands the
realized execution, while Sub and Fine measure whether that
understanding produces the correct continuation.}
\label{tab:diagnosis}
\centering
\fontsize{9}{10.8}\selectfont
\arrayrulecolor{tablerule}
\setlength{\tabcolsep}{0.48pt}
\renewcommand{\arraystretch}{1.08}
\begin{tabularx}{\linewidth}{@{}l*{8}{>{\centering\arraybackslash}X}@{}}
\toprule
\multirow{2}{*}{\tableheading{Model}}
& \multicolumn{4}{c}{\tableheading{Navigation}}
& \multicolumn{4}{c}{\tableheading{Operation}} \\
\cmidrule(lr){2-5}
\cmidrule(lr){6-9}
& \tableheading{Sub} & \tableheading{Fine} & \tableheading{Judge} & \tableheading{Cause}
& \tableheading{Sub} & \tableheading{Fine} & \tableheading{Judge} & \tableheading{Cause} \\
\midrule
GPT-5.5 & 72.7 & 39.4 & 52.8 & 39.8 & 65.4 & 41.1 & 57.7 & 43.2 \\
RoboBrain2.5-4B & 26.1 & 21.2 & 31.5 & 12.2 & 29.7 & 24.1 & 33.5 & 12.3 \\
RynnBrain-8B & 23.4 & 19.4 & 23.6 & 13.6 & 31.6 & 25.8 & 31.9 & 18.7 \\
InternVL3-8B & 21.8 & 17.7 & 48.3 & 26.2 & 32.1 & 26.8 & 40.1 & 17.9 \\
MiMo-Embodied-7B & 31.6 & 25.3 & 30.1 & 16.9 & 35.6 & 27.6 & 22.2 & 19.3 \\
Qwen2.5-VL-7B & 23.2 & 19.3 & 52.5 & 23.8 & 28.8 & 26.0 & 33.7 & 13.5 \\
Qwen3-VL-32B & 48.7 & 39.8 & 49.5 & 29.2 & 42.9 & 34.3 & 54.6 & 33.2 \\
\midrule
RynnBrain-SFT & 68.6 & 66.7 & 75.9 & 53.4 & 62.5 & 61.6 & 80.7 & 51.4 \\
RynnBrain-OPD & 70.0 & 68.0 & 79.0 & 57.0 & \textbf{65.4} & 61.0 & 80.0 & 54.0 \\
RynnBrain-GRPO & \textbf{73.0} & 69.2 & 74.2 & 51.5 & 63.6 & 62.1 & 76.3 & 49.6 \\
\rowcolor{tablehighlight}
\tablemethod{RynnBrain-NEPO} & 72.6 & \textbf{71.6} & \textbf{79.6} & \textbf{58.2}
& 64.9 & \textbf{63.7} & \textbf{81.2} & \textbf{54.6} \\

\rowcolor{tableablation}
- w/o Ent.-Gate & 69.9 & 68.5 & 73.1 & 50.8 & 63.1 & 61.4 & 75.0 & 48.7 \\

\bottomrule
\end{tabularx}
\end{table*}

Table~\ref{tab:diagnosis} shows that general VLMs can
occasionally distinguish success from failure, but their substantially
lower Cause and Fine scores indicate that outcome recognition alone does
not yield a usable correction. SFT substantially improves consistency
among the four fields, while NEPO provides the most balanced performance
across navigation and manipulation. It reaches Judge accuracies of
79.6\% and 81.2\%, Cause accuracies of 58.2\% and 54.6\%, and at least
63.7\% on all four recovery-planning scores. The model therefore learns
a relation between a failure cause and the way the interrupted subtask should be executed.

\paragraph{Transfer to an external planning benchmark.}
Because EmbodiedDiag-Bench follows the same output structure as the
training data, we additionally evaluate frozen checkpoints on
EB-ALFRED~\citep{yang2025embodiedbench}. No explicit failure prompt, code executor, VLA, or corrective
Harness is used. All models receive the same instruction, observation,
action space, history, and step limit. This experiment tests whether
failure-oriented training learns reusable task dependencies rather than
only the Judge--Cause--Sub--Fine response format.

\looseness=-1
As shown in Table~\ref{tab:eb_alfred_transfer}, Embodied-Corrector
improves over the base brain on all six subsets. Relative to SFT, the
additional gains of NEPO are most pronounced on Complex, Spatial, and
Long tasks, which require stronger prerequisite tracking and action
ordering. Controlled failure data therefore improves ordinary embodied
planning by teaching the model how task progress depends on previously
established conditions. The smaller, less structured gains on appearance-dependent cases show that correction training can reason over visual evidence but cannot replace information absent from the observation.

\subsection{Harness-Level Evaluation}
\label{sec:harness_eval}

\begin{wraptable}{r}{0.47\textwidth}
\vspace{-8pt}
\caption{Planning transfer to EB-ALFRED (\%). Frozen checkpoints receive the same inputs and step limit, without explicit failure prompts or a corrective Harness. Gains on Complex, Spatial, and Long tasks indicate reusable prerequisite tracking and action ordering. Com., Cplx., Vis., and Spat. denote Common, Complex, Visual, and Spatial.}
\label{tab:eb_alfred_transfer}
\centering
\fontsize{9}{10.8}\selectfont
\arrayrulecolor{tablerule}
\setlength{\tabcolsep}{0.55pt}
\renewcommand{\arraystretch}{1.08}
\begin{tabularx}{\linewidth}{@{}l*{6}{>{\centering\arraybackslash}X}@{}}
\toprule
\tableheading{Model} & \tableheading{Base} & \tableheading{Com.} & \tableheading{Cplx.} & \tableheading{Vis.} & \tableheading{Spat.} & \tableheading{Long} \\
\midrule
GPT-4o & 64 & 54 & 68 & 46 & 52 & 54 \\
Claude-3.7-Sonnet & 68 & \textbf{68} & 70 & \textbf{68} & 62 & \textbf{70} \\

InternVL3-8B & 20 & 14 & 14 & 12 & 0 & 2 \\
Ovis2-16B & 26 & 16 & 24 & 12 & 16 & 4 \\
Qwen3-VL-32B & 62 & 48 & 46 & 42 & 38 & 46 \\
Gemma-3-12B-IT & 32 & 26 & 38 & 22 & 20 & 12 \\
RoboBrain2.5-4B & 22 & 16 & 34 & 6 & 16 & 24 \\
MiMo-Embodied-7B & 20 & 16 & 24 & 18 & 20 & 18 \\
RynnBrain-8B & 30 & 28 & 44 & 26 & 24 & 28 \\
\midrule
RynnBrain-SFT & 72 & 48 & 62 & 54 & 56 & 52 \\
RynnBrain-OPD & 74 & 50 & 64 & 52 & 58 & 54 \\
RynnBrain-GRPO & 76 & 54 & 72 & 56 & 60 & 58 \\
\rowcolor{tablehighlight}
\tablemethod{RynnBrain-NEPO} & \textbf{78} & 56 & \textbf{74} & 60 & \textbf{64} & 62 \\
\rowcolor{tableablation}
- w/o Ent.-Gate & 74 & 52 & 68 & 54 & 58 & 56 \\

\bottomrule
\end{tabularx}
\end{wraptable}

Brain-level VQA accuracy does not establish that a correction remains
useful after workflow generation and physical execution. We therefore
evaluate the complete Harness under two mutually exclusive function
configurations. The first disables the VLA function and tests correction
through analytic robot functions and generated code. The second disables
all other robot functions and tests coordination with a frozen VLA. The
planning model, memory, execution interface, and budget remain fixed
within each comparison.

\paragraph{Code execution without the VLA.}
We remove $f_{\mathrm{VLA}}$ from the function library, so the
GPT code generator can invoke only analytic
functions for object detection, navigation, pose estimation, arm motion,
gripper control, and articulation. Each function returns its arguments,
result or exception, and available pre- and post-execution observations,
combining an ASPIRE-style execution trace with a CaP-X-style functional
robot interface
~\citep{lu2026aspire,fu2026capx}.

\begin{table*}[!htbp]
\caption{Code-Harness evaluation on OmniBehavior and LIBERO-PRO (\%). The VLA function is disabled in the code workflow. The GPT-5.5 code model, analytic robot functions, perception, and execution budget stay fixed when comparing corrective brains.}
\label{tab:code_execution}
\centering
\fontsize{9}{10.8}\selectfont
\arrayrulecolor{tablerule}
\setlength{\tabcolsep}{1.05pt}
\renewcommand{\arraystretch}{1.08}
\begin{tabularx}{\linewidth}{@{}l*{10}{>{\centering\arraybackslash}X}@{}}
\toprule
\tableheading{Model} & \multicolumn{4}{c}{\tableheading{OmniBehavior}} & \multicolumn{6}{c}{\tableheading{LIBERO-PRO}} \\
\cmidrule(lr){2-5}\cmidrule(lr){6-11}
& \tableheading{\shortstack{Rad.-\\Nav}} & \tableheading{\shortstack{Rad.-\\Op}} & \tableheading{\shortstack{Soda-\\Nav}} & \tableheading{\shortstack{Soda-\\Op}} & \tableheading{Obj.-P} & \tableheading{Obj.-T} & \tableheading{Goal-P} & \tableheading{Goal-T} & \tableheading{Spa.-P} & \tableheading{Spa.-T} \\
\midrule
OpenVLA & 4 & 0 & 2 & 0 & 0 & 0 & 0 & 0 & 0 & 0 \\
$\pi_0$ & 8 & 2 & 6 & 0 & 0 & 0 & 0 & 0 & 0 & 0 \\
$\pi_{0.5}$ & 20 & 8 & 18 & 6 & 17 & 1 & 38 & 0 & 20 & 1 \\
\midrule
GPT 5.5 & 76 & 44 & 80 & 60 & 61 & 58 & 72 & 77 & \textbf{58} & 66 \\
RynnBrain & 72 & 36 & 72 & 56 & 48 & 46 & 64 & 62 & 45 & 49 \\
\rowcolor{tablehighlight}
\tablemethod{RynnBrain-NEPO} & 76 & 40 & 80 & 64 & \textbf{69} & \textbf{67} & \textbf{82} & \textbf{85} & 56 & \textbf{70} \\
\bottomrule
\end{tabularx}
\end{table*}

We fix the code model GPT-5.5, function documentation,
perception module, tasks, seeds, and execution budget, and replace only the corrective brain. Embodied-Corrector outperforms the untrained brain under all ten conditions. Relative to GPT self-correction, it performs better in Libero-Pro. Because neither code generation nor the robot functions change, the result shows that a Judge--Cause--Sub--Fine response enables the same workflow model to revise the relevant function calls instead of repeating the original failed program.

\paragraph{Frozen-VLA execution.}
In the complementary setting, we evaluate HarnessVLA augmented with our Embodied-Corrector, while the execution library contains only $f_{\mathrm{VLA}}$. Analytic functions for object detection, navigation, pose control, or gripper control are disabled. Embodied-Corrector produces a state-conditioned subtask, fine-grained action text, and execution-error diagnosis and correction; GPT converts these outputs into HarnessVLA code that invokes a frozen $\pi_0$ or $\pi_{0.5}$ through \texttt{vla\_execute}. The None condition uses the same pipeline without the Corrector. VLA parameters, observations, action interface, and task split are unchanged, so the comparison isolates whether high-level correction and decomposition reduce the execution burden on an existing action model~\citep{zhang2026harnessvla,nasiriany2026robocasa365}.


\par
\ifdim\dimexpr\pagegoal-\pagetotal\relax<300pt
\newpage
\fi
\begin{wrapfigure}{r}{0.49\textwidth}
\vspace{-8pt}
\centering
\includegraphics[width=\linewidth]{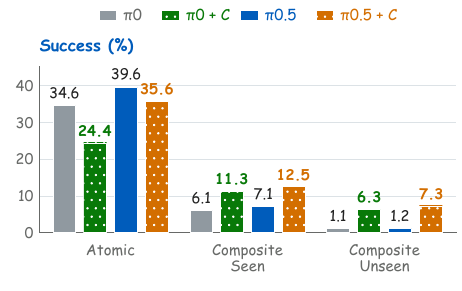}
\caption{Frozen-VLA success on RoboCasa365 (\%). Corrective planning improves composite tasks, while atomic success decreases.}
\label{fig:robocasa}
\end{wrapfigure}

The effect depends strongly on task structure. Atomic-Seen tasks already provide short instructions close to the frozen VLA's training distribution. Reformulating them can introduce an unnecessary language interface error, and atomic success consequently decreases for both policies. Composite-Seen and Composite-Unseen tasks instead require the VLA to map a long compositional instruction directly to actions. Embodied-Corrector converts the instruction into short, state-conditioned subtasks and substantially improves both groups. For $\pi_{0.5}$, Composite-Seen and Composite-Unseen success increases from 7.1\% and 1.2\% to 12.5\% and 7.3\%, respectively, producing an overall gain. The result shows that the brain does not replace low-level visuomotor control; it reduces the temporal and compositional complexity that a fixed VLA must solve at once. The atomic degradation identifies brain--VLA language alignment as a remaining system bottleneck.

\subsection{Post-Training Dynamics}
\label{sec:training_dynamics}

\begin{figure}[!htb]
\centering
\includegraphics[width=1\linewidth]{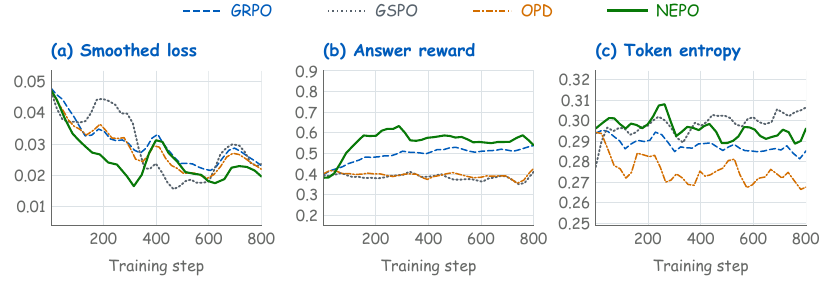}
\caption{VQA post-training dynamics from a shared SFT checkpoint. Curves report policy loss, answer-level reward, and sampled-token entropy under matched training settings. NEPO achieves the highest reward while avoiding the low-entropy regime of vanilla OPD.}
\label{fig:optimization}
\end{figure}

We next examine whether NEPO changes optimization behavior rather than only final benchmark accuracy. GRPO~\citep{shao2024deepseekmath}, GSPO~\citep{zheng2025gspo}, OPD, and NEPO start from the same evaluation-first SFT checkpoint and use the same multimodal VQA data, number of samples, response budget, seed schedule, and update count. Figure~\ref{fig:optimization} reports policy loss, answer-level VQA reward, and sampled-token entropy.

\begingroup
\widowpenalty=10000
Policy loss is non-monotonic for all methods and does not by itself explain their final separation. Vanilla OPD provides dense token-level guidance, but low-quality student prefixes create distribution mismatch between the student and privileged branches. The resulting noisy negative probability gaps gradually move OPD toward a lower-entropy regime. NEPO preserves positive privileged guidance while gating negative signals at uncertain positions. It consequently obtains the highest reward while avoiding the entropy contraction of vanilla OPD. The joint reward--entropy pattern shows that asymmetric gating improves the reliability of OPD supervision.
\par\endgroup

\subsection{Corrective Brain as a Subtask Reward}
\label{sec:reward_model_exp}

\begin{wraptable}{R}{0.49\textwidth}
\vspace{-8pt}
\caption{LIBERO success with terminal reward and optional frozen-brain subtask reward (\%). Corrective judgments provide intermediate feedback and chiefly improve Long tasks with delayed dependencies. Spa. and Obj. denote Spatial and Object.}
\label{tab:libero_reward_model}
\centering
\fontsize{9}{10.8}\selectfont
\arrayrulecolor{tablerule}
\setlength{\tabcolsep}{1.2pt}
\renewcommand{\arraystretch}{1.08}
\begin{tabularx}{\linewidth}{@{}ll*{5}{>{\centering\arraybackslash}X}@{}}
\toprule
\tableheading{VLA} & \tableheading{Brain} & \tableheading{Spa.} & \tableheading{Obj.} & \tableheading{Goal} & \tableheading{Long} & \tableheading{Avg.} \\
\midrule
\multirow{3}{*}{$\pi_0$} & None & \textbf{66} & 69 & \textbf{65} & 42 & 60.5 \\
 & RynnBrain-8B & 62 & 66 & 62 & 38 & 57.0 \\
\rowcolor{tablehighlight}
& \tablemethod{RynnBrain-NEPO} & 65 & \textbf{70} & 64 & \textbf{49} & \textbf{62.0} \\
\midrule
\multirow{3}{*}{$\pi_{0.5}$} & None & \textbf{86} & 78 & \textbf{79} & 56 & 74.8 \\
 & RynnBrain-8B & 82 & 75 & 75 & 52 & 71.0 \\
\rowcolor{tablehighlight}
& \tablemethod{RynnBrain-NEPO} & 85 & \textbf{79} & 77 & \textbf{63} & \textbf{76.0} \\
\bottomrule
\end{tabularx}
\end{wraptable}

We additionally freeze either the base brain or Embodied-Corrector and
use its binary subtask-completion judgment as an auxiliary reward for
VLA PPO~\citep{schulman2017ppo}. All conditions share the same policy initialization, terminal
environment reward, tasks, interaction budget, and optimization setup.
This experiment tests whether the learned judgment can shorten delayed
credit assignment rather than merely improve VQA classification.

The trained corrective brain primarily benefits Long tasks, whose
intermediate dependencies are difficult to localize from terminal reward
alone. On the shorter Spatial, Object, and Goal suites, subtask judgment
often duplicates information already contained in the terminal reward
and provides limited benefit. Inaccurate judgments from the base brain
can instead direct policy updates away from eventual task success. A
brain reward is therefore useful when it reliably exposes delayed task
structure, neutral when it repeats the environment signal, and harmful
when the two disagree.

\WFclear
\subsection{Real-Robot Validation}
\label{sec:piper_main}

We evaluate a Piper manipulator with an end-effector gripper and an RGB-D camera on five tabletop tasks: placing a Rubik's cube inside a blue plate; stacking the cube on a blue block; sorting an orange-colored fruit into an orange plate and a blue block into a blue plate, with other objects as distractors; moving the cube to the left of the blue block; and placing both designated fruits inside an open basket.

Each frozen backend is evaluated with and without Embodied-Corrector over 20 episodes per task, yielding 100 episodes per method and 400 overall. Correction uses observations and public execution traces to revise the next VLA instruction from the current physical state, preserving completed progress without resetting the scene. Detailed platform settings, success criteria, correction budgets, and representative execution frames remain in Appendix~\ref{app:piper}.

\begin{wrapfigure}{r}{0.49\textwidth}
\vspace{-8pt}
\centering
\includegraphics[width=\linewidth]{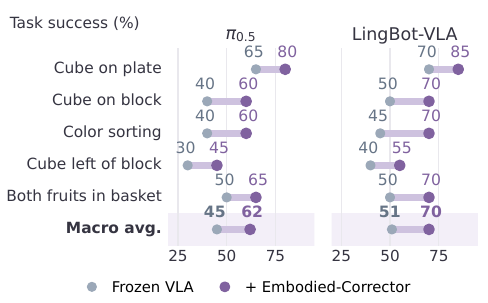}
\caption{Piper task success (\%; 20 episodes per task and method). Gray and purple markers show each frozen VLA without and with Embodied-Corrector. Macro averages weight the five tasks equally.}
\label{fig:piper_results}
\end{wrapfigure}

Figure~\ref{fig:piper_results} shows improvements on every task for both backends. Macro-average success rises from 45\% to 62\% for $\pi_{0.5}$ and from 51\% to 70\% for LingBot-VLA. LingBot-VLA color sorting gains 25 percentage points (45\% to 70\%); spatial rearrangement remains the hardest task, at 45\% and 55\% with correction.

The real-robot gains primarily stem from the correction mechanism rather than from strengthening the underlying VLA: it shifts failure handling from outcome-level retries to boundary-level cause localization and state-conditioned recovery. After each action chunk, the Corrector uses boundary observations and public execution traces to determine whether the current subtask is complete and to identify the observable failure cause; without resetting the scene, it then either advances or re-executes only the interrupted subtask, while rewriting the remaining goal into short, state-conditioned instructions that the frozen executor can reliably realize. This both prevents local execution deviations from propagating along task dependencies and reduces the compositional burden that a long instruction places on the frozen VLA at once.

\WFclear
\begin{figure}[!htbp]
\centering
\includegraphics[width=\linewidth]{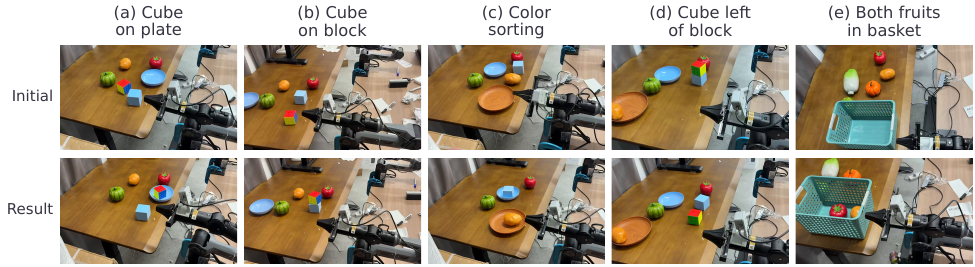}
\caption{Initial (top) and result (bottom) observations from the five Piper tasks, ordered as in the text. Result frames show the scene after the gripper releases the manipulated objects. Detailed execution sequences and success criteria are provided in Appendix~\ref{app:piper}.}
\label{fig:piper_initial_final}
\end{figure}

\FloatBarrier

\section{Conclusion and Limitations}
\label{sec:conclusion}

We presented Embodied-Corrector, viewing correction not as an auxiliary response to planning failure but as a necessary process for keeping subsequent plans grounded in the physical state. By aligning expected postconditions, observed outcomes, observable failure causes, and recovery decisions, it revises the task state before continuing, converting local deviations into specific recoveries rather than letting them propagate through task dependencies.
Real-robot evaluation currently covers five tabletop tasks; broader task and embodiment generalization remain to be evaluated. Observation noise and state uncertainty still complicate outcome judgment and recovery verification; future work will broaden real-world evaluation through uncertainty-aware postcondition assessment and safety-constrained recovery.

\bibliographystyle{acl_natbib}
\bibliography{iclr2027_conference}
\clearpage
\beginappendix

\section{Implementation Details}
\label{app:implementation}

\subsection{Data Composition}
\label{app:data_composition}

Training uses two sources: our controlled correction data and the Embodied-Reasoner corpus. The controlled data start from 300 RoboCasa365 tasks, containing 6,210 episodes and 2,078,151 raw rounds, and 50 OmniBehavior tasks, containing 328 episodes and 358,608 raw rounds~\citep{nasiriany2026robocasa365,li2024behavior1k}. We inject controlled perturbations at subtask boundaries and construct Judge, Cause, Sub, and
Fine supervision from the resulting executions. After filtering for
effective perturbations, complete observations, observable failures, and
duplicate trajectories, the corpus contains 1,583,014 valid rounds from
148,415 rollouts. Source trajectories are split before perturbation
branches are generated, preventing examples with the same successful
prefix from appearing in both the training set and EmbodiedDiag-Bench.

Evaluation-first SFT mixes the controlled correction data with general
embodied-planning data from Embodied-Reasoner. The controlled data teach
execution judgment, failure-cause analysis, and corrective planning,
whereas Embodied-Reasoner preserves general task decomposition,
object-relation reasoning, and action-dependency knowledge
~\citep{zhang2025embodiedreasoner}.

The second stage also operates on fixed multimodal VQA examples and
contains no robot interaction or environment rollout. Its data consist
of 28,871 correction responses sampled from the student distribution
and 3,003 Embodied-Reasoner planning examples, producing 31,874 records
in total. 
All post-training methods share the same data, initialization, prompts,
and response budget; they differ only in their reward objective and in
whether privileged evaluation and entropy gating are used.

\subsection{Unified Harness Evaluation Protocol}
\label{app:unified_harness}

\paragraph{Common workflow.}
All execution experiments use the same
``plan--execute functions--correct--revise workflow''
loop. A fixed closed-source GPT-5.5 model performs both high-level task
planning and workflow generation. It receives the task goal, current
observation, within-episode execution history, previous correction,
robot-function documentation, and retrieved memory, and produces the
next workflow.

Multimodal memory is provided only to GPT-5.5 and is not an input to
Embodied-Corrector. Working memory maintains the current observation,
completed steps, latest correction, and remaining task. Episodic memory
stores environment-validated correction cases, while semantic memory
stores cause--repair rules consolidated from repeated successful cases.
To keep evaluation trials independent, formal experiments retain only
within-episode working memory and disable cross-episode episodic
retrieval, memory writing, and semantic-rule updates.

After a robot-function call or VLA action chunk returns, the Harness
records pre- and post-execution images, function arguments, return values
or exceptions, and observable changes in robot and object state.
Following the execution-trace-based correction mechanism of ASPIRE,
this multimodal evidence is passed to Embodied-Corrector
~\citep{lu2026aspire}. The corrector receives neither retrieved memory
nor hidden simulator state. In a single inference, it produces Judge,
Cause, Sub, and Fine: Judge determines whether the current subtask was
completed, Cause identifies a failure supported by observable evidence,
Sub specifies the next subtask, and Fine describes how that subtask
should be executed.

The structured correction is then returned to GPT-5.5. After success,
GPT-5.5 preserves completed progress and continues the remaining task.
After failure, it revises the subsequent workflow according to Cause,
Sub, and Fine. GPT-5.5 receives only the structured correction record,
not the private reasoning of Embodied-Corrector.
OmniBehavior, LIBERO-PRO, and RoboCasa365 execution experiments use simulation and benchmark-defined success predicates. Piper hardware uses the independent evaluator and protocol in Appendix~\ref{app:piper}.

\paragraph{Code Harness with the VLA function disabled.}
In the first configuration, $f_{\mathrm{VLA}}$ is removed from the function library. GPT-5.5 can compose only analytic robot functions, such as
\texttt{detect\_object},
\texttt{navigate\_to},
\texttt{estimate\_pose},
\texttt{move\_to\_pose},
\texttt{control\_gripper}, and
\texttt{open\_joint}. These functions share a common argument, return, and exception interface and expose multimodal execution traces for correction. This configuration combines a CaP-X-style functional robot
interface with ASPIRE-style execution feedback
~\citep{fu2026capx,lu2026aspire}.

OmniBehavior separately evaluates navigation and manipulation success for the radio and soda-can tasks, giving four metrics:
Radio-Navigation, Radio-Operation, Soda-Navigation, and Soda-Operation. Each task--metric condition is evaluated from 25 distinct random initializations. Each reported cell is therefore the mean binary success over 25 complete trials.

The first three rows of Table~\ref{tab:code_execution}
(OpenVLA, $\pi_0$, and $\pi_{0.5}$) are results reported by CaP-X under its original evaluation protocols and were not rerun in our Harness; they are therefore excluded from the 25- and 100-trial settings described above~\citep{fu2026capx}.

For the remaining three rows, all compared conditions use the same GPT-5.5 model, robot-function documentation, perception module, task set, initial-state distribution, and execution budget. Only the correction component is changed among GPT-5.5 self-correction, RynnBrain-8B, and Embodied-Corrector. Cross-task memory writing and skill-library updates are disabled, so any change in success measures whether the correction helps the same planner revise subsequent function calls.

\paragraph{HarnessVLA execution with a frozen VLA as the only available robot function.}
All conditions use HarnessVLA for robot execution. In this configuration,
the function library exposed to the GPT code generator contains only
\texttt{vla\_execute}, backed by a frozen $\pi_0$ or $\pi_{0.5}$. Analytic
functions for object detection, navigation, pose control, gripper
control, and articulation are disabled.

At fixed action-chunk boundaries, Embodied-Corrector receives the overall
task, the previous stage instruction, and the pre- and post-execution
observations. In brain-assisted conditions, it generates a complete
Judge--Cause--Sub--Fine record in one inference. This record contains
subtask decomposition, fine-grained textual action planning, and
execution-error diagnosis and correction. These structured outputs are
then passed to GPT, which generates executable HarnessVLA code. Under the
frozen-VLA-only configuration, the generated code can invoke only
\texttt{vla\_execute}; in configurations with a larger function library,
the same code-generation mechanism can also invoke analytic tools or VLA
functions.

The generated code is executed by HarnessVLA. The frozen VLA receives
image observations and robot state through the unchanged HarnessVLA
interface, together with the language instruction specified by the
generated execution call. VLA parameters, observation preprocessing,
action space, and task horizon remain unchanged across all conditions.

Evaluation is performed through the same HarnessVLA execution pipeline.
It covers 18 Atomic-Seen, 16 Composite-Seen, and 16 Composite-Unseen
RoboCasa365 tasks
~\citep{nasiriany2026robocasa365}. Each brain-assisted condition runs 50
independent trials per task, yielding 2,500 episodes for each
brain--VLA pair: 900 Atomic-Seen, 800 Composite-Seen, and 800
Composite-Unseen episodes.

\section{Detailed protocol for the perturbations}
\label{app:table1_perturbations}

\paragraph{Shared intervention protocol.}
Each example starts from a retained successful trajectory at the state immediately before a subtask boundary. We
clone that state and hold the task instruction, successful prefix, intended object and target, and all non-selected
execution factors fixed. We then intervene on exactly one of the seven factors below and replay only the affected
subtask. An intervention is retained only if the expected postcondition is violated, the consequence is observable in
the synchronized public observation or execution record, and the branch remains recoverable. Ineffective
interventions, branches with multiple changed factors, and failures that can be identified only from hidden simulator
state are discarded. Recovery always begins from the realized failed state rather than from a reset. Because scene
geometry, joint ranges, and skill duration differ across tasks, perturbation validity is defined by the resulting
predicate change rather than by one global displacement, angle, or timestep range. This construction makes the factor
name a causal intervention label, not a free-form description assigned after a naturally occurring failure.

\paragraph{Navigation.}
This intervention changes the executed navigation goal, the desired terminal heading, or both, while leaving the
semantic destination and manipulation request unchanged. It is applied at a navigation boundary before the affected
navigation skill is replayed. The resulting stop can lie outside the interaction region or face the workspace from an
incorrect approach direction, so the target-region or pose postcondition fails even though locomotion itself may
finish normally. A matched correction estimates a new reachable goal and heading from the current observation,
navigates there, verifies that the interaction region is in range, and then resumes the blocked manipulation. This
factor tests whether the brain can distinguish successful motion from task-relevant arrival and repair a geometric
navigation error instead of repeating the same command.

\paragraph{Reachability.}
Reachability keeps the intended destination fixed but makes the original waypoint or terminal stop pose unusable. The
intervention either blocks that pose or invalidates it under the planner's collision and kinematic feasibility checks.
Unlike Navigation, which perturbs where or how the agent is asked to arrive, Reachability perturbs whether the nominal
route or stop configuration can be executed. The observed consequence is a no-path result, an unreachable target, or
a stop from which the next end-effector goal cannot be reached. Recovery resamples a collision-free intermediate
waypoint or stop pose, verifies planning feasibility, and returns to the original subtask. The diagnostic requirement
is to propose an alternative feasible route or base configuration rather than misclassifying the failure as object
absence or a grasp error.

\paragraph{Grasp pose.}
This intervention changes the end-effector approach direction, contact height, or grasp orientation for the specified
object while preserving object identity and the requested downstream task. The perturbed grasp can miss the object,
contact an unstable region, or initially contact and then slip. Offline construction uses pose, contact, and attachment
signals to confirm the label, while the deployed brain must infer the failure from public images and the execution
record. The correction re-localizes the object, recomputes a reachable pre-grasp and grasp pose, approaches without
collision, closes the gripper, and verifies attachment before lifting or transporting. This factor evaluates spatial
diagnosis at the object--gripper interface and whether the recovery restores a stable grasp rather than merely issuing
another close command.

\paragraph{Timing.}
Timing changes when the gripper open or close command occurs relative to approach, contact, lift, transport, or
release. The geometric target and grasp pose remain fixed, isolating temporal coordination from spatial accuracy. An
early close can prevent the gripper from enclosing the object; a late close can begin after contact has been lost; an
early open can release during transport; and a late open can leave placement incomplete. The common observable
outcomes are an unsecured object or a drop. Recovery first restores the relevant pre-contact or pre-release relation,
then issues the gripper command at the correct phase and verifies attachment or release before continuing. This factor
tests whether the brain attributes failure to action order and synchronization rather than unnecessarily changing the
object, goal, or grasp geometry.

\paragraph{Container.}
The Container intervention changes the articulation state of a receptacle such as a drawer, cabinet, or door by
leaving it only partially open or partially closed at the subtask boundary. The container identity and requested
content operation remain unchanged, but the opening, clearance, or required joint-state predicate is absent. The next
reach, placement, retrieval, or closure can therefore be infeasible even if its own command is well formed. A matched
correction re-localizes the articulated part, completes the required open or close motion past the task-defined
postcondition, verifies the resulting space or state, and resumes the blocked content operation. This factor tests
whether the brain recognizes articulation as a physical precondition rather than repeatedly repairing the downstream
grasp or placement.

\paragraph{Prerequisite.}
This intervention removes one dependency from the executed task prefix while keeping the blocked subtask request
unchanged. Typical dependencies include reaching the workspace before manipulation, opening a container before
insertion or retrieval, holding an object before transport, or clearing a destination before placement. The requested
subtask may be syntactically valid but is infeasible in the realized state because its first required predicate is
false. Recovery identifies the earliest unmet dependency, executes and verifies it, and then returns to the original
blocked subtask without discarding the successful prefix. This factor directly measures causal task-graph reasoning:
the correct response must repair an earlier condition, not paraphrase or locally retry the final command.

\paragraph{Target.}
Target intervention changes the grounded same-class instance or the executable reference while preserving the task
instruction and skill type. For example, a command intended for one bowl can be bound to another visible bowl, so a
technically successful grasp or placement still changes the wrong entity. The failure is accepted only when instance
identity and the task relation disagree and the mismatch is observable in the retained evidence. Recovery resolves the
instruction against the current scene, re-grounds the specified instance and destination, undoes or safely terminates
the wrong-object branch when necessary, and replans the blocked operation. This factor tests reference resolution and
instance-level grounding, separating semantic target errors from failures of navigation, reachability, or control.

\section{Corrective Harness}
\label{sec:harness}


\paragraph{Problem formulation.}
Given a long-horizon task goal $g$, at execution round $t$, a fixed
closed-source GPT model receives the current observation $o_t$, execution
history $H_t$, retrieved memory $\mathcal{R}(M_t)$, the previous correction
record $e_{t-1}$, and the robot-function documentation
$\mathcal{D}_{\mathcal{F}}$. It outputs an executable workflow $w_t$.
After the workflow is executed, the Harness returns the new observation
$o_{t+1}$ and execution trace $\tau_t$. Embodied-Corrector then produces
a structured correction record from the resulting execution evidence:
\begin{equation}
w_t =
\operatorname{GPT}
\!\left(
g,o_t,H_t,\mathcal{R}(M_t),e_{t-1},
\mathcal{D}_{\mathcal{F}}
\right),
e_t=(d_t,c_t,r_t)
=
\mathcal{C}_{\theta}
\!\left(
g,w_t,o_t,o_{t+1},\tau_t
\right).
\label{eq:corrective_harness}
\end{equation}
Here, $w_t$ denotes workflow code composed of robot-function calls, and $\mathcal{D}_{\mathcal{F}}$ describes callable functions for object detection, navigation, pose control, gripper control, articulation, and a frozen VLA. The outputs $d_t$, $c_t$, and $r_t$ denote the subtask completion judgment, the failure cause supported by execution evidence,
and the recovery proposal from the current state, respectively.
Retrieved memory is provided only to the GPT model, whereas
Embodied-Corrector reasons only from the task, the executed workflow,
the pre- and post-execution observations, and the returned trace.
The GPT model then conditions on $e_t$ in the next round to continue or
revise the workflow while preserving valid prior progress.

\paragraph{Workflow.}
As illustrated in Figure~\ref{fig:harness_architecture}, the closed-source planning model first retrieves task-relevant context from the multimodal memory shown on the left. Working memory maintains the current task state, the latest execution outcome, and the unfinished subtask. Episodic memory stores environment-validated successful corrections, such as a case in which a skipped grasp must be inserted before transport. Semantic memory consolidates reusable task rules, such as requiring object attachment before transport and verifying that attachment before proceeding. These memories are provided only to the closed-source planning and workflow-code model, not to Embodied-Corrector. When retrieved experience conflicts with the current observation, the current physical evidence takes precedence.

The closed-source planning model combines the task goal, current state, and retrieved memory to directly generate a workflow over robot functions. Following the function-based robot interface of CaP-X, the function set $\mathcal{F}$ may include
\texttt{detect\_object}, \texttt{navigate\_to}, \texttt{approach},
\texttt{close\_gripper}, \texttt{lift}, \texttt{open\_joint}, and
\texttt{verify\_attachment}
~\citep{fu2026capx}.
Following Harness VLA, a frozen VLA is exposed through the same function library as
\texttt{vla\_execute(instruction, observation, robot\_state)}, which realizes local contact-rich skills such as grasping, placing, and articulation
~\citep{zhang2026harnessvla}.
The VLA is therefore not a separate execution system outside the workflow. It is one robot function, alongside navigation, pose control, gripper control, and other analytic functions.

After a robot function or VLA function is executed, the fixed execution Harness returns the pre- and post-execution observations, function arguments, return value or exception, and observable changes in robot and object state. Following the execution-trace-based correction mechanism of ASPIRE, this multimodal evidence is passed to Embodied-Corrector
~\citep{lu2026aspire}.
The corrector neither retrieves memory nor directly generates executable workflow code. Instead, it determines whether the requested subtask was actually completed, identifies a failure cause supported by the observations, and proposes the task condition or recovery steps that must be restored. In the LIBERO-PRO orange-juice example in the figure, the failed program attempts placement with an empty gripper while the carton remains on the floor. The corrector therefore marks placement as incomplete, identifies the omitted pickup prerequisite, and proposes locating and grasping the carton, verifying attachment, and placing it in the basket. The lower strip shows frames from the failed rollout followed by frames from the regenerated rollout, rather than one uninterrupted trajectory.

The correction record is then returned to the same closed-source planning model. After success, the planner preserves completed progress and continues the remaining workflow. After failure, it combines the recovery proposal with retrieved memory to regenerate the required function calls—for example, target localization, approach, gripper closure, lifting, and attachment verification before resuming placement. The new physical outcome is evaluated again, forming a loop of
``memory retrieval--workflow generation--function execution--execution correction--workflow revision.'' Working memory is updated after every round. A correction enters episodic memory only after the environment confirms that it restored task progress, and repeated compatible successes may later be consolidated into semantic rules. During formal evaluation, the closed-source planner, robot functions, frozen VLA, perception interface, and execution budget are held fixed, so the experiment measures whether Embodied-Corrector can restore task progress by changing the subsequent workflow.

\section{Public Robot Tool Interfaces for Closed-Loop Correction}

\subsection{Overview}
At the beginning of each subtask, the code-generation model receives a complete list of tool interfaces, including function signatures, parameter meanings, return values, coordinate frame conventions, and major failure conditions. The model is not allowed to send undefined robot control commands directly; instead, it generates Python programs composed solely of these public tools. This design decouples high-level reasoning from low-level execution: the ``brain'' is responsible for determining whether a subtask is completed, analyzing failure reasons, and generating fine-grained recovery plans; the code model translates recovery plans into calls for perception, geometric computation, grasp planning, and motion control.

The tool interfaces also serve as the adaptation boundary between simulation and real hardware. The simulation and real-robot backends may use different camera drivers, segmentation services, inverse kinematics solvers, and arm controllers, but they must preserve identical tool names, parameter semantics, coordinate frames, and return structures. Currently, the LIBERO/ASPIRE backend provides 24 public functions, among which point\_prompt\_molmo is registered only when the Molmo service is available. The code-generation model may define auxiliary functions local to the task, but it cannot bypass these interfaces to access private simulator states.

Environment success determination is not part of the tool interfaces. The generated program can only read public observations and robot states; it cannot access rewards, task-completion flags, ground-truth object poses, or internal contact states of the simulator. After the program terminates, an independent environment evaluator computes success or failure based on task predicates and returns the result to the Harness. Therefore, the model's own completion declarations do not directly affect the experimental results.

\subsection{Public Tool Interface Summary}
The following public tools are available to the code-generation model. The interface covers visual perception, 3D geometry, grasp planning, and robot control. All state-changing operations must be executed through the motion-control tools, while environment rewards and task-completion predicates remain hidden from the generated program.

\paragraph{Observation}
\begin{itemize}
\item \textbf{get\_observation()} – Returns a multimodal observation dictionary containing RGB, depth, intrinsics, and camera-to-world transforms from the main and wrist cameras, along with robot joint angles, end-effector pose, and gripper state. This is the sole entry point for obtaining the current environment state.
\end{itemize}

\paragraph{Vision}
\begin{itemize}
\item \textbf{segment\_sam3\_text\_prompt(rgb, text\_prompt)} – Returns a list of mask candidates (each with binary mask, bounding box, and confidence) for the text-described instance; returns an empty list if no target is detected.
\item \textbf{segment\_sam3\_point\_prompt(rgb, point\_coords)} – Segments the instance at the given pixel coordinate, useful when text-based segmentation is unstable or disambiguation is needed.
\item \textbf{point\_prompt\_molmo(image, text\_prompt)} – Returns the pixel coordinates of a text-described target using a vision-language model; registered only when the Molmo service is available.
\end{itemize}

\paragraph{Grasp Planning}
\begin{itemize}
\item \textbf{plan\_grasp(depth, intrinsics, segmentation)} – Generates grasp candidates (transform matrix and confidence) from a depth image and target mask using Contact-GraspNet. Returned poses are in the camera frame and must be transformed to the world frame; reports failure if no candidate.
\item \textbf{plan\_grasp\_from\_point\_clouds(pc\_full, pc\_segment)} – Generates grasp candidates from the full scene point cloud and target object point cloud, suitable for multi-view fusion.
\item \textbf{select\_top\_down\_grasp(grasps, scores, cam\_to\_world, vertical\_threshold=0.8)} – Transforms candidates to the world frame, filters for near-vertical grasps, and returns the highest-scoring valid candidate.
\end{itemize}

\paragraph{Geometry}
\begin{itemize}
\item \textbf{depth\_to\_point\_cloud(depth, intrinsics)} – Back‑projects the entire depth map into a point cloud in the camera frame using the pinhole model.
\item \textbf{mask\_to\_world\_points(mask, depth, intrinsics, extrinsics)} – Back‑projects valid depth pixels within the mask to 3D and transforms them to the world frame.
\item \textbf{pixel\_to\_world\_point(u, v, z, intrinsics, extrinsics)} – Converts a single pixel and its depth to a 3D world point, useful for turning vision model pointing into robot targets.
\item \textbf{get\_oriented\_bounding\_box\_from\_3d\_points(points)} – Computes the oriented bounding box (center, extent, rotation matrix) of the target point cloud.
\item \textbf{interpolate\_segment(p1, p2, step=0.03)} – Generates evenly spaced waypoints between two endpoints, used for staged approach, lifting, transport, or pushing.
\item \textbf{normalize\_vector(v)} – Normalises a direction vector; near‑zero vectors are left unchanged to avoid numerical issues.
\end{itemize}

\paragraph{Point Cloud Processing}
\begin{itemize}
\item \textbf{subsample\_point\_cloud(pc, max\_points=10000)} – Downsamples the point cloud to reduce time and memory overhead.
\item \textbf{filter\_noise(points, colors=None)} – Removes outliers using density‑based clustering; preserves colour correspondence if provided.
\end{itemize}

\paragraph{Transformations}
\begin{itemize}
\item \textbf{transform\_points(points, transform\_matrix)} – Applies a \(4\times4\) homogeneous transform to a point set or organised point cloud.
\item \textbf{rotation\_matrix\_to\_quaternion(R)} – Converts a \(3\times3\) rotation matrix to a unit wxyz quaternion.
\item \textbf{decompose\_transform(T)} – Decomposes a \(4\times4\) homogeneous grasp transform into position and wxyz quaternion for control interfaces.
\end{itemize}

\paragraph{Motion Planning and Control}
\begin{itemize}
\item \textbf{solve\_ik(position, quaternion\_wxyz)} – Solves inverse kinematics for the Franka arm (7‑DOF). Falls back to predefined poses if the target is unreachable; reports error if all fail.
\item \textbf{move\_to\_joints(joints)} – Blocking move to the specified joint configuration; returns after motion completes.
\item \textbf{goto\_pose(position, quaternion\_wxyz, z\_approach=0.0)} – Wraps IK and joint motion. When \texttt{z\_approach} is set, moves to a point above the target then descends.
\end{itemize}

\paragraph{State Recovery}
\begin{itemize}
\item \textbf{goto\_home\_joint\_position()} – Returns the arm to a predefined home configuration. Should not be called while carrying an object.
\end{itemize}

\paragraph{Gripper Control}
\begin{itemize}
\item \textbf{open\_gripper()} – Fully opens the gripper.
\item \textbf{close\_gripper()} – Fully closes the gripper. Closing alone does not imply successful grasping; subsequent observations must verify stable lifting.
\end{itemize}

\subsection{Typical Tool Calling Workflow}
For a pick‑and‑place subtask, the generated program first calls get\_observation to acquire images and camera parameters, then uses text or point prompts to segment the target object. The target mask is converted to a world‑frame point cloud via mask\_to\_world\_points, which is used either to estimate the object centre or to generate grasp candidates. Subsequently, the program calls select\_top\_down\_grasp, decompose\_transform, and solve\_ik to obtain an executable pose, then executes the grasp through open\_gripper, goto\_pose, close\_gripper, and a lifting motion. After placement, the program calls get\_observation again to acquire evidence of the outcome. Only when the external environment evaluator confirms that the task predicate holds does the Harness mark the subtask as successful.

\subsection{Tool Call Logging and Error Correction}
Every tool call is written into a structured trajectory that includes at least the call order, function name, parameter summary, execution time, result summary, and corresponding visual keyframes. The correction model receives the task instruction, the failed program, the public observations, and this execution trajectory, and outputs four structured fields: whether the subtask succeeded, the observable failure reason, the next subtask, and a fine‑grained recovery plan. The code model then generates a revised program based on this record. In this way, the correction results reported in the paper are grounded in verifiable environmental evidence, rather than in the model's subjective judgment of its own program success.

\section{Correction Cases}
\label{app:representative_harness_cases}

The following examples show how our \textbf{Embodied-Corrector} resolves representative execution failures in LIBERO-PRO and RoboCasa365. 

\subsection{LIBERO-PRO: correction guides program regeneration}

\paragraph{L1, Prerequisite: pick up the orange juice before placement.}
\textbf{Task.} Pick the orange juice and place it in the basket.
\textbf{Failed execution.} The generated program starts from placement even though the orange-juice carton has not
been grasped. The empty gripper moves to the basket and opens, so neither the object state nor the task predicate changes.
\textbf{Correction.} Embodied-Corrector marks the current subtask as unsuccessful, identifies the omitted pickup as the
first causal error, selects ``localize and grasp the orange juice'' as the next subtask, and specifies grasp, attachment
verification, lift, and basket placement in the fine-grained plan. The code generator uses this record to restore the
missing dependency instead of refining the already invalid placement motion.
\textbf{Correct execution.} The regenerated program localizes and grasps the carton, verifies that it moves with the
gripper, transports it over the basket opening, and releases it inside. The paired rollout figure contrasts the empty
placement attempt with the corrected placement and the environment-verified success state.

\begin{figure}[htbp]
  \centering
  \includegraphics[width=0.99\linewidth]{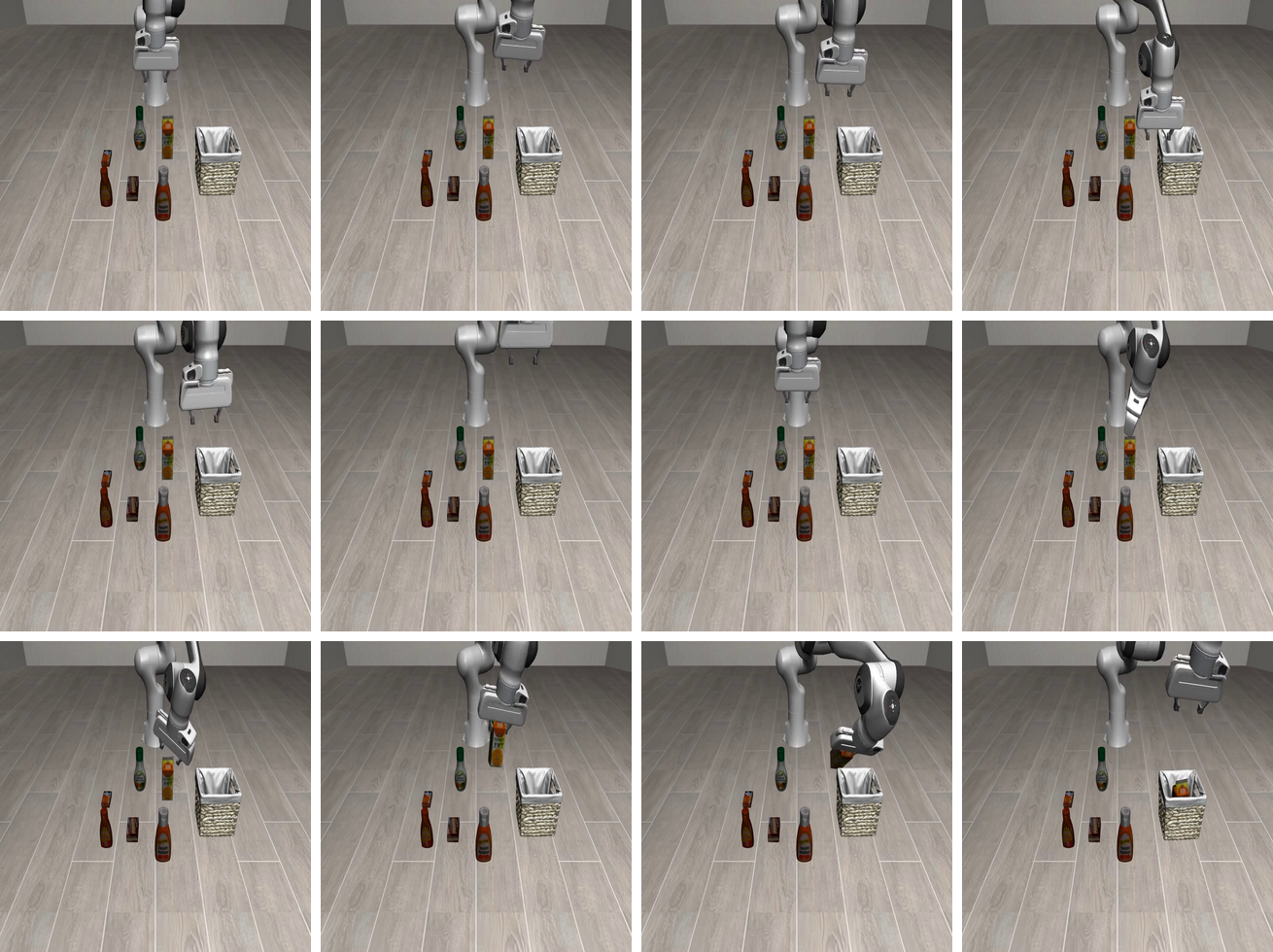}
  \caption{Prerequisite correction in LIBERO-PRO case L1. Twelve raw observations contrast an empty-gripper placement attempt with recovery of the missing pickup step. The corrected program grasps the orange-juice carton, verifies attachment, and places it in the basket.}
  \label{fig:libero_orange_case}
\end{figure}

\paragraph{L2, Target: resolve the requested bowl before control.}
\textbf{Task.} Pick the black bowl that is \emph{not between} the plate and the ramekin and place it on the plate.
\textbf{Failed execution.} The generated program admits a dark non-bowl region into its candidate set and plans around
that false referent. Although the arm executes plausible motions, acting on the wrong instance cannot satisfy the
requested spatial relation.
\textbf{Correction.} Embodied-Corrector attributes the failure to target selection, retains the placement objective,
and asks the code generator to ground the plate, ramekin, and true bowl instances before applying ``not between.'' The
fine-grained plan explicitly rejects non-bowl masks and the bowl lying on the plate--ramekin relation.
\textbf{Correct execution.} The corrected program selects the other black bowl, grasps and lifts it, clears the nearby
ramekin, and releases it over the plate center. The final spatial relation is accepted by the environment.

\begin{figure}[htbp]
  \centering
  \includegraphics[width=0.99\linewidth]{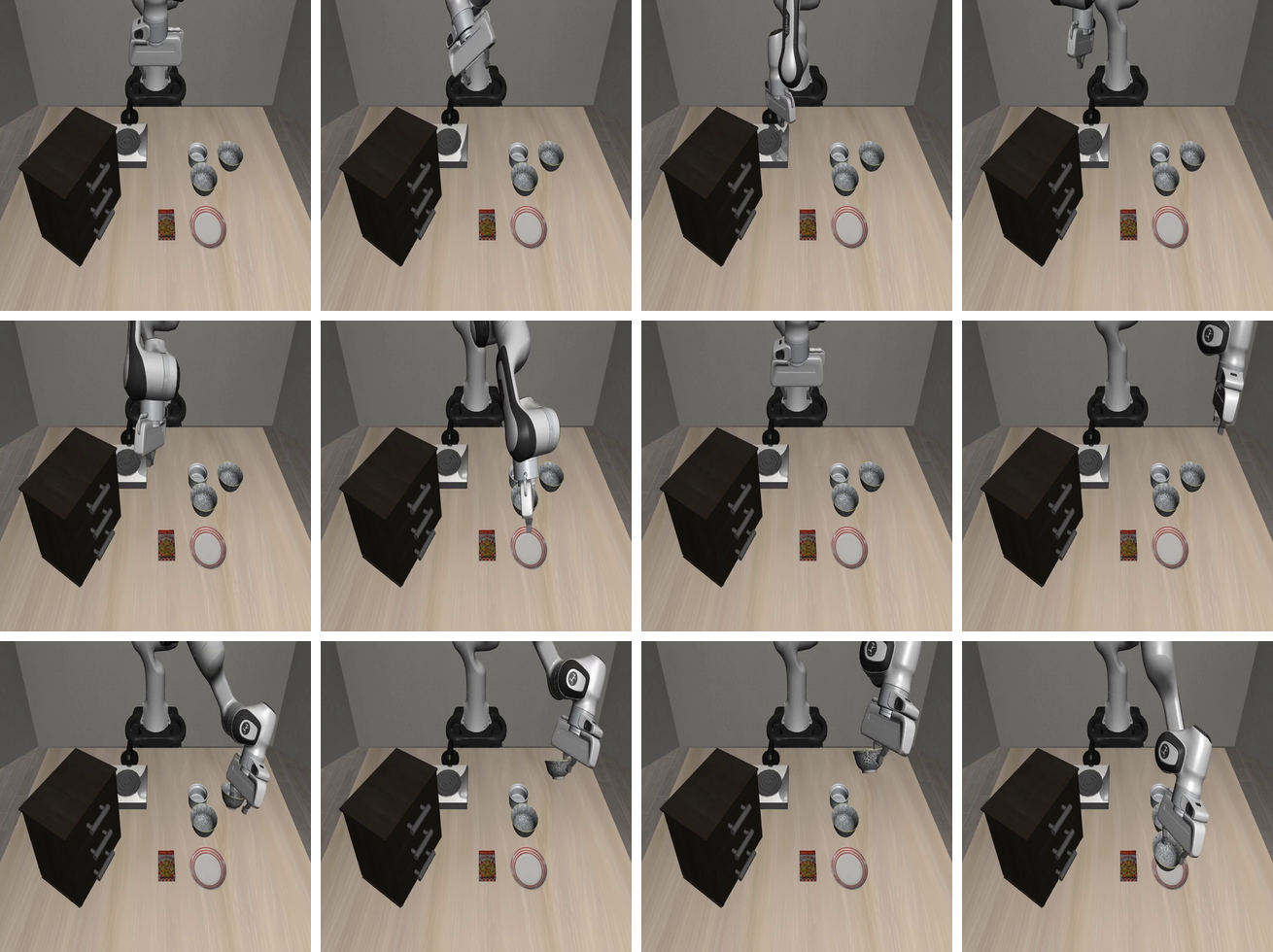}
  \caption{Target correction in LIBERO-PRO case L2. Twelve raw observations contrast false-target execution with re-grounding of the requested black bowl. The corrected program excludes non-bowl regions, applies the spatial constraint, and verifies placement on the plate.}
  \label{fig:libero_spatial_case}
\end{figure}



\subsection{RoboCasa365: correction revises the instruction to a frozen VLA}

In RoboCasa365, control returns to Embodied-Corrector after each VLA action chunk, while the frozen $\pi_{0.5}$ continues
to generate low-level actions. Unlike the paired LIBERO-PRO rollouts, each RoboCasa365 example is one continuous rollout.
Each figure shows twelve chronological observations spanning unsuccessful grasps, corrective instructions, and the
terminal successful state. These are raw observations from the same rollout; complete videos provide the temporal evidence.

\paragraph{R1, High and offset grasp: align the avocado.}
\textbf{Task.} Pick the avocado from the sink and place it on the plate located on the counter.
\textbf{Failed execution.} The rollout reaches over the sink but repeatedly closes without enclosing the avocado. The
object remains in the basin, exposing coupled errors in descent and lateral alignment.
\textbf{Correction.} Embodied-Corrector uses the unsuccessful status, boundary observations, and the history of empty
grasp attempts to change the next subtask to side approach and output a directional alignment instruction. As the view
changes, the corrector requests small left or right adjustments and only then restores the close-and-lift action. This
is an observation-conditioned correction while the VLA remains frozen.
\textbf{Correct execution.} The corrected instruction brings the gripper into the basin, restores lateral alignment,
and leads to the avocado leaving the sink and satisfying the environment's object--plate predicate.

\begin{figure}[htbp]
  \centering
  \includegraphics[width=0.99\linewidth]{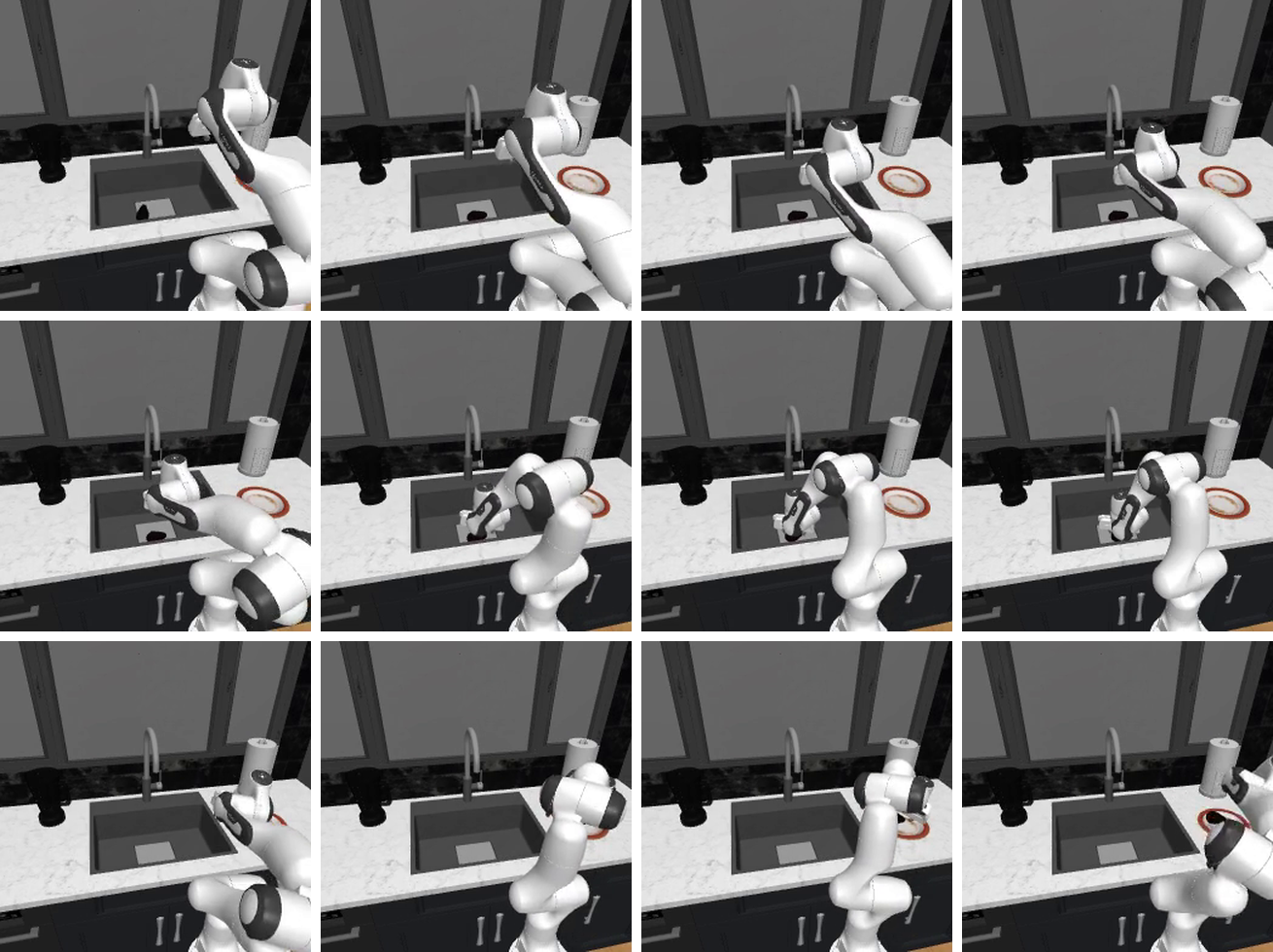}
  \caption{High and offset grasp correction in RoboCasa365 case R1. Twelve raw observations from one continuous rollout show failed avocado grasps followed by directional re-alignment. With the VLA frozen, corrected instructions restore pickup and successful placement on the plate.}
  \label{fig:robocasa_avocado_case}
\end{figure}

\paragraph{R2, Offset and late grasp: reposition before contacting the onion.}
\textbf{Task.} Pick the onion from the sink and place it on the plate located on the counter.
\textbf{Failed execution.} The first pickup instruction contains directional motion, but the contact timing is wrong:
the gripper closes too late, knocks the onion, and leaves it in the sink. Repeating closure without changing the approach
does not recover the object.
\textbf{Correction.} The corrector receives the directional plan, boundary observations, and
the earlier action history showing the unsuccessful grasp. It keeps the pickup subtask but makes recovery executable by coordinating forward, lateral,
and downward repositioning before closure. Later updates continue to adapt the approach as the view changes, rather than
repeating an unchanged close command.
\textbf{Correct execution.} The corrected approach encloses the onion, carries it out of the sink, and satisfies the
requested object--plate relation. RoboCasa365 verifies successful termination from its public task predicate.

\begin{figure}[htbp]
  \centering
  \includegraphics[width=0.99\linewidth]{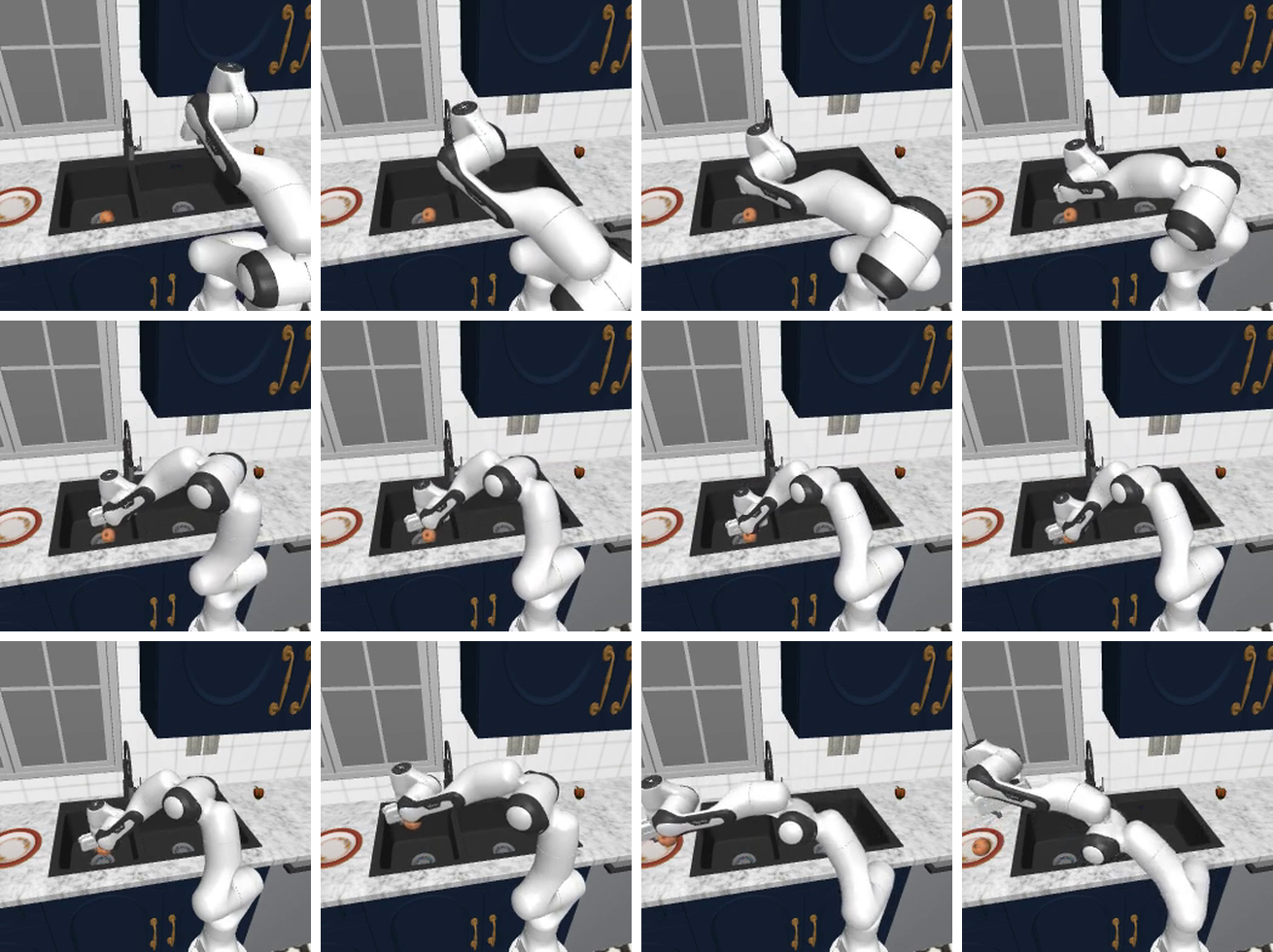}
  \caption{Offset and late-contact correction in RoboCasa365 case R2. Twelve raw observations from one continuous rollout show the onion left in the sink after late closure. Corrected instructions coordinate re-approach and closure before successful plate placement with the frozen VLA.}
  \label{fig:robocasa_onion_case}
\end{figure}

Across these cases, successful correction depends on replacing the failed execution premise: restore a prerequisite,
re-ground the target, or rebuild a stable grasp by changing approach height, lateral alignment, and contact timing.
Embodied-Corrector supplies this structured change, while the downstream code generator or frozen VLA executes it and the
environment verifies the final result.

\FloatBarrier

\section{Additional Evaluation Protocols}
\label{app:additional_protocols}

\subsection{EmbodiedDiag-Bench}
\label{app:embodieddiag}
EmbodiedDiag-Bench evaluates whether a model can understand failures from execution evidence and generate a recovery plan. It directly evaluates four structured fields and does not require physical robot control.

Each input contains the task goal, current subtask, execution history, pre- and post-execution observations, action parameters, function returns, and exception information. The model generates Judge, Cause, Sub, and Fine in that order. Judge is scored with binary accuracy. Cause identifies the failed entity, physical relation, and observable outcome; correctness requires the predicted components to agree with the reference. Sub identifies the recovery direction and target subtask, including continuation, repetition, recovery of an earlier prerequisite, or termination. Fine is matched by executable postconditions, necessary actions, target parameters, action order, and result verification rather than exact string match.

We score the four fields independently; joint correctness requires all fields to be correct. Open-ended Fine outputs that cannot be judged automatically are evaluated by two independent annotators using a shared rubric. The test set is generated only from held-out source trajectories; the train/test split is completed before perturbation generation, so successful execution prefixes are not shared with training.

\subsection{EmbodiedBench/EB-ALFRED}
\label{app:ebalfred}
EmbodiedBench/EB-ALFRED evaluates transfer to ordinary embodied planning. It provides no explicit failure prompt and does not use an online correction loop, a corrective Harness, a VLA backend, or an additional code executor.

All models receive the same task instruction, visual observations, action space, execution history, maximum execution steps, decoding settings, maximum response length, and random seeds. Inference starts from frozen checkpoints without test-time parameter updates or human feedback.

We report success rates on Base, Common, Complex, Visual, Spatial, and Long. The benchmark environment predicate determines episode success. We first average within each task and then macro-average across tasks; Table~\ref{tab:eb_alfred_transfer} reports the resulting success rates. This experiment measures ordinary planning and tests whether controlled failure data improves prerequisite tracking, action ordering, and task dependencies.

\section{Piper Real-Robot Experiments}
\label{app:piper}

We evaluate Embodied-Corrector on five Piper tabletop tasks using two frozen VLA backends, $\pi_{0.5}$~\citep{physicalintelligence2025pi05} and LingBot-VLA~\citep{wu2026pragmatic}, each with and without correction.

\begin{figure}[!htbp]
\centering
\includegraphics[width=\textwidth]{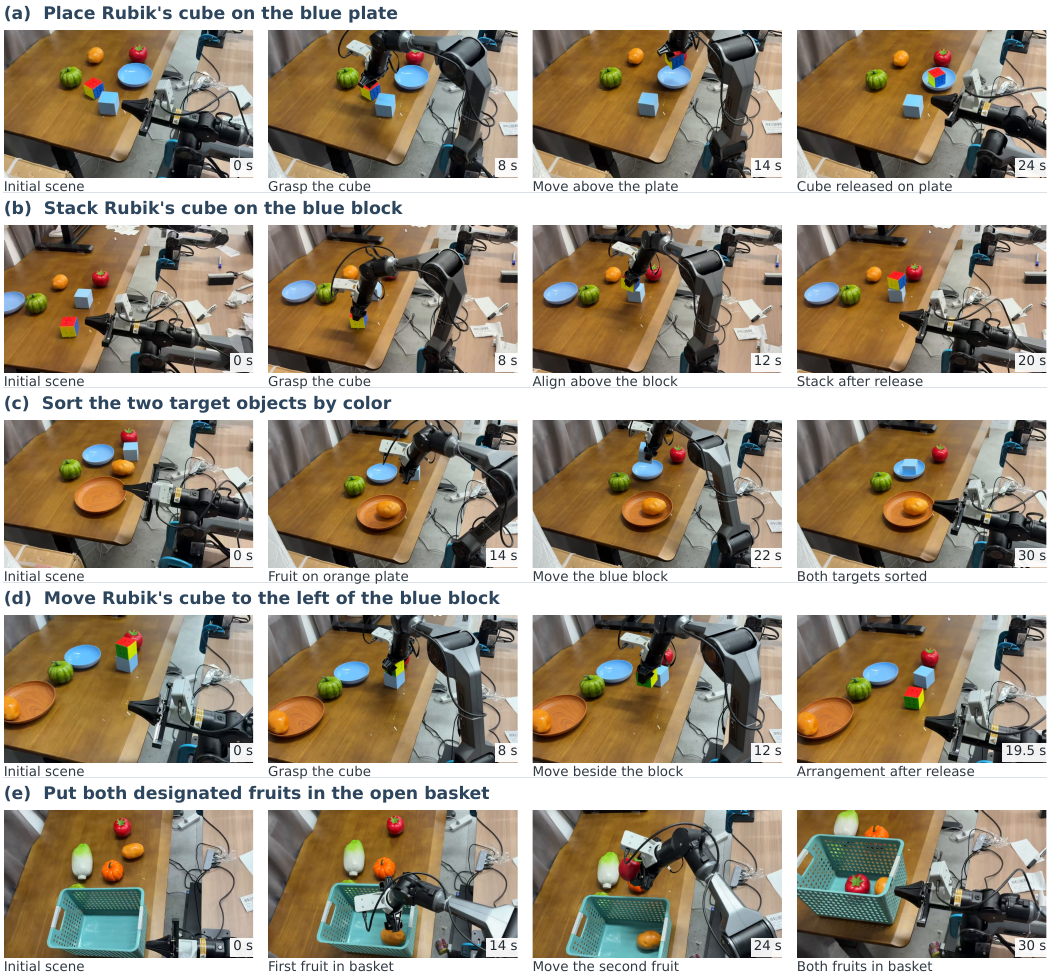}
\caption{Representative Piper executions for the five real-robot tasks. Each row shows four chronological frames from one video, read from left to right. Action labels describe visible execution stages; timestamps are relative to the start of the corresponding video. The color-sorting task targets the orange-colored fruit and blue block, and the basket task requires both designated fruits. Frames illustrate physical execution; aggregate success rates are reported in Figure~\ref{fig:piper_results}.}
\label{fig:piper_tasks}
\end{figure}

\subsection{Real-Robot Platform}
\label{app:piper_platform}
The platform comprises a Piper manipulator, an end-effector gripper, and an RGB-D camera. The frozen VLA generates an action chunk from the instruction and current visual observation. For each executed chunk, the system records pre- and post-execution RGB-D observations, robot state, the active instruction, execution outcome, exceptions, and execution time.

Embodied-Corrector receives the goal, current subtask, execution history, boundary observations, and public execution trace. Its four output fields are \textit{Judge}, which assesses subtask completion; \textit{Cause}, which describes observable outcomes and failure evidence; \textit{Sub}, which specifies the next subtask; and \textit{Fine}, which provides recovery actions and verification steps. The correction record becomes the next VLA instruction, preserving completed progress.

The corrector cannot access hidden environment states, task rewards, private contact states, or human-provided success labels. VLA parameters remain fixed throughout evaluation.

\subsection{Tasks and Success Criteria}
\label{app:piper_tasks}
The tasks cover target selection, stacking, color sorting, spatial rearrangement, and sequential container placement. Figure~\ref{fig:piper_tasks} shows the actual objects and executions.

\paragraph{Place Rubik's cube on the plate.}
The robot selects the Rubik's cube among the tabletop objects and places it inside the blue plate. Success requires the released cube to rest within the plate, rather than in the gripper or outside the plate.

\paragraph{Stack Rubik's cube on the blue block.}
The robot places the Rubik's cube on the blue block. Success requires the stack to remain stable after release throughout the evaluation observation period.

\paragraph{Sort the two target objects by color.}
The robot places the orange-colored fruit in the orange plate and the blue block in the blue plate; other objects are distractors. Both targets must be released inside their matching plates, without omissions or color mismatches.

\paragraph{Move Rubik's cube to the left of the blue block.}
Let $\mathbf{p}_{R}$ and $\mathbf{p}_{B}$ denote the tabletop centers of the Rubik's cube and blue block, with $x$ increasing from left to right in the evaluation frame. Success requires
\begin{equation}
    x_R < x_B, \qquad
    5\,\mathrm{cm} < \lVert\mathbf{p}_{R}-\mathbf{p}_{B}\rVert_2 < 10\,\mathrm{cm}.
\end{equation}
Both distance bounds are strict.

\paragraph{Put both designated fruits in the open basket.}
The robot places the orange-colored fruit and red fruit into the open basket. Both must lie fully inside after release; an object held by the gripper, on the rim, or outside does not satisfy the criterion.

\subsection{Evaluation Protocol}
\label{app:piper_protocol}
Each method runs $N=20$ independent episodes per task: 100 per method and 400 overall. Each episode starts from a predefined physical arrangement and an initial VLA action chunk.

With Embodied-Corrector, the system reacquires RGB-D observations and robot state after each chunk and assesses subtask completion. Completed subtasks advance while preserving progress; incomplete subtasks trigger recovery instructions. Recovery continues from the actual robot and object configuration, without resetting the scene within an episode.

Each episode allows at most $K=10$ correction attempts, counted as recovery instructions for incomplete subtasks. Judgments that advance completed subtasks do not consume this allowance. Success requires meeting the task criterion within the allowance; exhaustion without completion is failure. Collision risk, workspace violations, risk of object damage, or a manual safety stop immediately terminate the episode as unsuccessful.

\subsection{Metrics and Results}
\label{app:piper_results}
An independent task evaluator determines final success using Sec.~\ref{app:piper_tasks}; model-generated ``success'' text does not determine outcomes. Task success rates use 20 episodes each, and the macro average weights the five tasks equally.

The protocol additionally records first-pass success (completion without correction), correction attempts, total execution time, VLA action chunks, corrector invocations, correction latency, and safety terminations. Invocations include assessments of completed subtasks and therefore differ from correction attempts. Figure~\ref{fig:piper_results} reports final success rates.

Embodied-Corrector improves all five tasks for both backends. Macro success rises from 45\% to 62\% for $\pi_{0.5}$ and from 51\% to 70\% for LingBot-VLA: gains of 17 and 19 percentage points. The largest task-specific gain is LingBot-VLA color sorting, from 45\% to 70\%. Spatial rearrangement remains the hardest task for both augmented systems (45\% and 55\%), indicating that precise placement remains challenging.

\end{document}